\documentclass[journal]{IEEEtai}

\usepackage{cite}
\usepackage{amsmath,amssymb,amsfonts}
\usepackage{algorithmic}
\usepackage{graphicx}
\usepackage{textcomp}
\usepackage{times}
\usepackage{epsfig}
\usepackage{bbm}
\usepackage{array}
\usepackage{subfig}
\graphicspath{{./}}
\usepackage{comment}
\usepackage{float}
\usepackage{multicol}
\usepackage{ragged2e}
\usepackage{enumerate}
\usepackage{multirow}
\usepackage{mathabx}
\usepackage{algorithmic}
\usepackage{soul}
\usepackage[ruled,vlined]{algorithm2e}
\usepackage[pagebackref=true,breaklinks=true,colorlinks,bookmarks=false]{hyperref}

\def\BibTeX{{\rm B\kern-.05em{\sc i\kern-.025em b}\kern-.08em
    T\kern-.1667em\lower.7ex\hbox{E}\kern-.125emX}}

\begin{document}

\title{SKID: Self-Supervised Learning for Knee Injury Diagnosis from MRI Data}
\author{Siladittya Manna, Saumik Bhattacharya, Umapada Pal, \IEEEmembership{Senior Member, IEEE}

\thanks{Siladittya Manna is with the Computer Vision and Patter Recognition Unit, Indian Statistical Institute Kolkata, India (email: siladittya\_r@isical.ac.in)}
\thanks{Saumik Bhattacharya is with the Department of Electronics and Electrical Communication Engineering, Indian Institute of Technology Kharagpur, India (email: saumik@ece.iitkgp.ac.in)}
\thanks{Umapada Pal is with the Computer Vision and Patter Recognition Unit, Indian Statistical Institute Kolkata, India (email: umapada@isical.ac.in)}
}

\markboth{Journal of IEEE Transactions on Artificial Intelligence, Vol. 00, No. 0, Month 2020}
{S. Manna \MakeLowercase{\textit{et al.}}: SKID: Self-Supervised Learning for Knee Injury Diagnosis from MRI Data}

\maketitle

\begin{abstract}
In medical image analysis, the cost of acquiring high-quality data and their annotation by experts is a barrier in many medical applications. Most of the techniques used are based on supervised learning framework and need a large amount of annotated data to achieve satisfactory performance. As an alternative, in this paper, we propose a self-supervised learning (SSL) approach to learn the spatial anatomical representations from the frames of magnetic resonance (MR) video clips for the diagnosis of knee medical conditions. The pretext model learns meaningful spatial context-invariant representations. The downstream task in our paper is a class imbalanced multi-label classification. Different experiments show that the features learnt by the pretext model provide competitive performance in the downstream task. Moreover, the efficiency and reliability of the proposed pretext model in learning representations of minority classes without applying any strategy towards imbalance in the dataset can be seen from the results. To the best of our knowledge, this work is the first work of its kind in showing the effectiveness and reliability of self-supervised learning algorithms in class imbalanced multi-label classification tasks on MR videos.\\
   \indent The code for evaluation of the proposed work is available at \\ \href{https://github.com/sadimanna/skid}{\fontfamily{qcr}\selectfont https://github.com/sadimanna/skid}.
\end{abstract}

\begin{IEEEImpStatement}
Self-Supervised learning has emerged as one of the best tools for representation learning without using human-annotated data. % Similar to transfer learning, Self-supervised learning also consists of two parts, pretext and downstream. However, unlike transfer learning, the pretext task is not performed using ground truth labels of the dataset used. Rather, in self-supervised learning, a pretext task is designed which helps in learning generalized robust representations from the data. Furthermore, 
%The representations learnt in the self-supervised pretext task provide a better starting point in the loss landscape in the downstream task, as the dataset used in the pretext and downstream task is the same. %It has been observed in several research studies that self-supervised pre-trained weights provide better results than ImageNet pre-trained weights on several tasks. 
In this paper, we design a self-supervised framework that can effectively classify different knee injuries from MR scans. Apart from handling this multi-label classification task, we also highlighted the limitations of the existing geometrical transformation prediction-based pretext tasks, which will help us to understand the latent behaviours of the SSL algorithms. 

\end{IEEEImpStatement}


\section{Introduction}
\label{sec:introduction}
Computer Vision tasks like object detection, segmentation or tracking have reached near human precision and reliability with the application of deep learning techniques \cite{alexnet,vggnet,effnet,sqandextnet,densenet,resnet}. These deep learning techniques are mainly supervised in nature and need a huge amount of annotated data for learning proper generalization. In absence of enough data, the deep learning models often suffer from the overfitting problem. Collection as well as annotation of such huge amount of data is expensive and time consuming. In case of medical image data, the collection of data requires some expensive apparatus and the annotation needs to be done with the help of experienced medical personnel.

The drawbacks of the supervised learning algorithms as mentioned above have led researchers to devise techniques which enable algorithms to learn meaningful representations of the data without human annotated labels. Such techniques fall under the paradigm of self-supervised learning. Similar to Transfer learning, Self-supervised learning primarily consists of two phases: (1) Pretext (also called Pre-training) and (2) Downstream. The primary objective of the pretext task is to train the encoder to learn representations which can then be transferred to the downstream or target task, which is generally a classification or segmentation problem using the representations learnt from the pretext phase. However, unlike transfer learning, the pretext task is not a supervised one. Rather, in self-supervised learning the pretext task is designed specifically to solve a task with pseudo-labels instead of the ground truth labels. The primary objective of the pretext task is to learn generalized and robust representations from the data without using the ground truth labels. This helps in providing a better initialization point in the loss landscape for the downstream task. In transfer learning scenario, the dataset used in pretext and downstream tasks are different. Thus, the weights learnt in transfer learning pretext task are hierarchically adapted to the dataset used in the pretext task. If the downstream task dataset is small and different from the pretext dataset, then it presents a risk of overfitting and the co-adaptation between the hierarchical features is destroyed, as small dataset provide less samples for reconstruction of the same \cite{howtr}. However, in case of self-supervised learning, the dataset used in the pretext and downstream task is the same. Hence, the weights learnt in the pretext task are adapted to the downstream dataset and the hierarchical co-adaptation of the features are not destroyed.

\indent
In this paper, we propose a self-supervised deep learning architecture called SKID for medical diagnosis of MR videos. %The pretext task in our work is based on the principle of jigsaw puzzle solving strategy. 
With rigorous experimental evidence, we have shown that the proposed method (SKID) learns spatial context-invariant features from the dataset without prior knowledge of the imbalanced distributions and the inter-dependency of the classes. In our work, the downstream task is a multi-label classification problem and its objective is to predict one or more anomalies present in the MR videos. \\
\indent
Main contributions of our work are as follows:
\begin{itemize}
    \item An effective algorithm is proposed in learning meaningful features without prior knowledge of the underlying class distribution and inter-class dependency. This is validated through different experiments.
    \item A novel network architecture, named SKID is proposed for Jigsaw Puzzle Solving as the pretext task. This novel architecture learns useful information and discards redundant information by using a bottleneck technique in the Dimension Reduction block.
    \item To the best of our knowledge, this work is the first of its kind to show the effectiveness and reliability of self-supervised learning algorithm in multi-label classification task on MR video.
\end{itemize}

The rest of the paper is organized as follows: Section \ref{sec:litsurvey} discusses some of the related works done in the recent years. Section \ref{sec:methodology} describes the methodology including the algorithms and model architectures of the pretext and downstream tasks in detail. In Section \ref{sec:exp}, we describe the experimental details and examine the results obtained. Section \ref{sec:effects_GTP_pre} explains the effects of low level signals that hinder the representation learning process in self-supervised learning. Section \ref{sec: more_abl} details about the ablation studies done under this work. Finally, Section \ref{sec:conclusion} concludes the paper.

\section{Literature Survey}
\label{sec:litsurvey}

 Data in computer vision applications like medical image analysis are often hard to annotate because of the lack of expert personnel. Furthermore, the lack of data in such applications makes it difficult to train deep neural networks from scratch. Transfer learning, semi-supervised learning, weakly-supervised learning, domain adaptation are some examples of such techniques which do not require a large volume of annotated data for training \cite{surveytransl, surveysemisup, surveysemisup2, surveyweaksup}. Recently, advances have been made to develop techniques \cite{shuffleandlearn, temporder, oddoneout, cliporder, sssptemporder, imgcolor, rotnet, videorotnet, contextpred} which can learn meaningful representations of the underlying data without human annotations. From this point of view, these techniques can be classified as a type of unsupervised learning. In contemporary literature, these techniques are termed as self-supervised learning. These techniques consist of two phases- pretext and downstream. %The self-supervised learning models learn meaningful features by means of solving various tasks called \textit{Pretext} tasks. 
 Several types of pretext tasks such as image inpainting \cite{contextenc},
%, jigsaw puzzle solving \cite{noroozi, damagedjig, videojig, iterreorg},
temporal order correction or verification \cite{shuffleandlearn, temporder, oddoneout, sssptemporder},
clip order prediction \cite{cliporder},
image coloring \cite{imgcolor},
geometric transformation prediction \cite{rotnet, videorotnet},
relative patch prediction based on context-aware features \cite{contextpred},
etc. have been proposed. %Another interesting approach for learning representations from videos in the pre-training phase is cycle-consistency learning via forward-backward tracking \cite{twiaa}. In \cite{sslvos}, reconstruction of target frame from the reference frame of a video is used for the pretext task. Among many others, self-supervised scene flow estimation \cite{sslsfe}, self-supervised data augmentation \cite{sslreid}, self-supervised disparity prediction \cite{ssldisp} are also used as pretext tasks.
The features learnt by the pretext model serve as the \textit{pre-trained} features in the downstream task and hence, the pretext tasks are carefully designed such that the pretext model can extract meaningful representations from the data.% by means of solving the objectives defined in the pretext task. \\

\indent In \cite{rotnet}, representations are learnt by predicting the angle by which an image has been rotated. In \cite{contextenc}, the model is made to generate the missing contents from the image. This requires the model to understand the surroundings of the missing part and learn visual semantic structures of the data as well. Among other techniques, image coloring \cite{imgcolor} maps the grayscale input to quantized colour value in the CIE \textit{Lab} colour scale and treats the problem as multinomial classification. Methods described in \cite{shuffleandlearn, temporder, oddoneout, cliporder, sssptemporder} learn representations from a video in a self-supervised manner by exploiting the information stored in the temporal dimension. Shuffling video frames helps the model learn the change in the spatial features over time and helps in capturing both spatial and temporal features. Relative patch prediction method like \cite{contextpred} divides an image into different patches and predicts the position of a given patch to learn representations from the data. This allows the model to learn context-aware features. The common feature among all the aforementioned self-supervised algorithms is that the pretext tasks used in these algorithms are hand-crafted. 

\indent
 Jiao et al. \cite{jnoble} proposed a combination of temporal order prediction and geometric transformation prediction on ultrasound videos for representation learning. The learnt representations were used for standard plane detection as the downstream task. However, to the best of our knowledge none of the above algorithms has explored the ability of such self-supervised learning algorithms in learning features from a class imbalanced multi-label dataset. Recently, Sowrirajan et al. \cite{mococxr} applied Momentum Contrast \cite{moco} method to learn meaningful transferable representations from Chest X-Ray images and have achieved better performance than supervised learning algorithms. 
 
 \indent
 In the aforementioned algorithms \cite{rotnet, imgcolor, shuffleandlearn, temporder, oddoneout, cliporder, sssptemporder, contextpred, jnoble}, the representations learnt by the model are based on natural images. In spite of development of several self-supervised learning algorithms for natural images or videos, application of such algorithms in medical image analysis remains limited. Medical data need to be handled in a different way than natural image data because of its nature. Medical data are grayscale and contain features which have small spatio-temporal extent. Furthermore, the dimensions of medical data are often large because of the need to capture minute details for proper diagnosis. Thus, a model needs to be capable of learning fine-grained representations from the data. To achieve this objective, we propose a novel architecture, named SKID, which learns useful representations by predicting the order of arrangements in a jigsaw puzzle task.

\section{Methodology}
\label{sec:methodology}
\indent
Self-supervised learning in medical data analysis is particularly challenging as the known pretext tasks extract irrelevant features frequently. We have discussed this in detail in Section \ref{sec:effects_GTP_pre}. The goal of our work is to extract meaningful spatial representations from the MR data with minimal low level signal contamination like empty regions, image boundary pixels, etc. We achieve this goal by devising a novel architecture (SKID), which predicts the arrangement of the patches obtained %into which a single frame of the \textcolor{red}{MR data}s has been divided 
using PREPFRAM (Algorithm \ref{alg:alg1}) described in Sec. \ref{subsubsec:pretext_algorithm}. In the following subsections we discuss about the pretext task algorithm and model architecture followed by a detailed discussion on the downstream ensembling strategy for diagnosing MR videos. % in a class imbalanced multi-label classification task scenario.

\subsection{Pretext Task}
\label{subsec:pretext_task}
\indent
The pretext task in this paper is based on Jigsaw Puzzle solving strategy \cite{noroozi, damagedjig, videojig, iterreorg}. However, we would like to term our pretext task as Jumbled Patch Arrangement prediction task as the proposed model predicts the order of arrangement of the patches that each frame is divided into. In the following subsections, we discuss about the algorithm and the model architecture used in the pretext task.

\subsubsection{Algorithm}
\label{subsubsec:pretext_algorithm}
\indent
 In this learning strategy, a frame is chosen randomly according to an uniform distribution from a magnetic resonance video. This frame is then divided into $N$ square patches. Each patch is of dimensions $\lfloor \frac{L}{\sqrt{N}}\rfloor \times \lfloor \frac{L}{\sqrt{N}}\rfloor$, where $L$ is the dimension of each square frame. Dividing a frame into $N$ patches gives $N!$ ways to jumble the $N$ patches. If we consider $N = 9$, the number of ways the patches can be jumbled is $9! = 3,62,880$. Let, $\mathcal{J}$ be the set of all possible arrangements of the patches. Solving a classification task with such large number of classes would require huge computational time and resources. To circumvent the issue, we choose 1000 elements from the set $\mathcal{J}$ randomly according to an uniform distribution and form the set $A$.

\begin{figure}[ht]
    \centering
    \begin{tabular}{cc}
            \subfloat[]{\includegraphics[width = 1.6in]{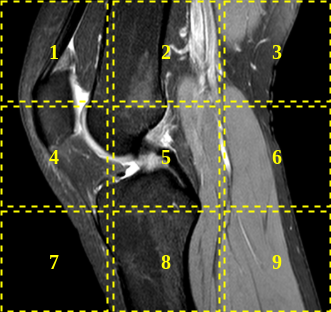} \label{fig:egimga}} &
            \subfloat[]{\includegraphics[width = 1.6in]{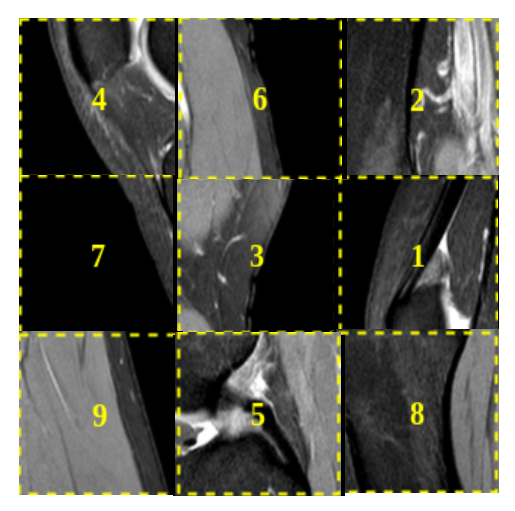} \label{fig:egimgb}}\\
    \end{tabular}
    \caption{(a) Example of an image showing the position of the patches in order. (b) Image showing the patches after being arranged in a randomly chosen order. The image is best visible in 200\% scale.}
    \label{fig:ExampleImages}
\end{figure}

\indent
Data augmentation was applied to increase the diversity of the data and make the model robust. The data augmentations applied include horizontal and vertical shift, rotation and scaling. Horizontal and vertical shift values were randomly selected from $[-\lfloor0.1 L_p \rfloor,\lfloor0.1 L_p \rfloor]$ and rotation values were selected from the range $[-15^{\circ},15^{\circ}]$. The scale factor was however chosen from a finite set $\mathcal{S} = \{1,1.2\}$. Here, $L_p$ denotes the length of each side of a square patch and $L_p \neq \frac{L}{\sqrt{N}}$. We also applied additive white Gaussian noise with mean 0 and variance 0.01 to the image patches before feeding them to the model.\\
\indent To obtain the jumbled patches and the pretext labels, we apply the algorithm PREPFRAM (Algorithm \ref{alg:alg1}) to the selected frames. First, a frame is divided into $\sqrt{N} \times \sqrt{N}$ parts ($\mathcal{P}'$) with each part having dimensions $\lfloor \frac{L}{\sqrt{N}} \rfloor \times \lfloor \frac{L}{\sqrt{N}} \rfloor$ as shown in Fig. \ref{fig:egimga}. Augmentation $g$ is applied to each of the $N$ parts to generate augmented partition $\mathcal{P}_{g}$. Then, for each $\mathcal{P}_{g}$, we sample a point ($ref_{x}$,$ref_{y}$) from the range $[0,\lfloor \frac{L}{\sqrt{N}} \rfloor - 64]$ randomly according to a uniform distribution $\mathcal{U}$. Using this point ($ref_{x}$,$ref_{y}$) as origin, we then crop a $64 \times 64$ region ($\mathcal{P}_{64}$) from the partition. %Augmenting the original patch before cropping a $64 \times 64$ region decreases the possibility of unwanted edges being formed towards the boundary region of each patch and consequently prevents the model from learning unwanted low-level features. 
In our experiments, we set $\lfloor \frac{L}{\sqrt{N}} \rfloor = 85$. Finally, the arrangement $\mathcal{A}_{\tau}$ is applied on the patches $\mathcal{P}_{A}$ to get the jumbled patches $\mathcal{P}_{\tau}$ as shown in Fig. \ref{fig:egimgb}.

\begin{algorithm}
\SetAlgoLined
\KwResult{$\mathcal{P}_A$ : Jumbled patches from a frame $\mathcal{F}$}
 \textbf{Given}\\
 $A$ : Set of 1000 arrangements\\
 $L$  : Length of each side of square frame\\
 $N$ : Number of patches each frame is divided into\\
 $g(\cdot)$ : Random augmentation applied on patches\\
 $\mathcal{U}_z [a,b]$ : $z$ is a sample drawn from a uniform distribution with range $[a,b]$\\
 $\mathcal{A}_{\tau}(\cdot)$ : An arrangement drawn from a uniform distribution on the set $A$\\
 \textbf{Initialize} \\
 $\mathcal{F}$ = a frame from a MR video\\
 $\mathcal{P}_A = \{\,\}$\\
 $\mathcal{L'} = \lfloor \frac{L}{\sqrt{N}} \rfloor$\\
 $row = col = ref_{x} = ref_{y} = 0$\\
 \For {i = 1 : 9}{
 $row = \lfloor \frac{i}{\sqrt{N}} \rfloor$ \\
 $row' = row + 1$\\
 $col = i \mod \sqrt{N}$\\
 $col' = col + 1$\\
 $\mathcal{P'} =  \mathcal{F}[row. \mathcal{L'}  : row'.\mathcal{L'} , col. \mathcal{L'} : col'. \mathcal{L'}]$\\
 $\mathcal{P}_{g} = g(\mathcal{P'})$\\
 $ref_{x} = \mathcal{U}_x [0,\mathcal{L'}-64]$\\
 $ref_{y} = \mathcal{U}_y [0,\mathcal{L'}-64]$\\
 $\mathcal{P}_{64} = \mathcal{P}_{g}[ref_{x} : ref_{x}+64, ref_{y} : ref_{y}+64]$\\
 $\mathcal{P}_A = \mathcal{P}_A \bigcup \mathcal{P}_{64}$\\
 $\mathcal{P}_{\tau} = \mathcal{A}_{\tau}(\mathcal{P}_A)$\\
 }
 \caption{PREPFRAM : How to prepare each frame for training}
 \label{alg:alg1}
\end{algorithm}

\subsubsection{Model Architecture}
\label{subsubsec:pretext_model_arch}

\begin{figure*}[ht]
\begin{minipage}{\textwidth}
    \centering
    \includegraphics[width =\linewidth]{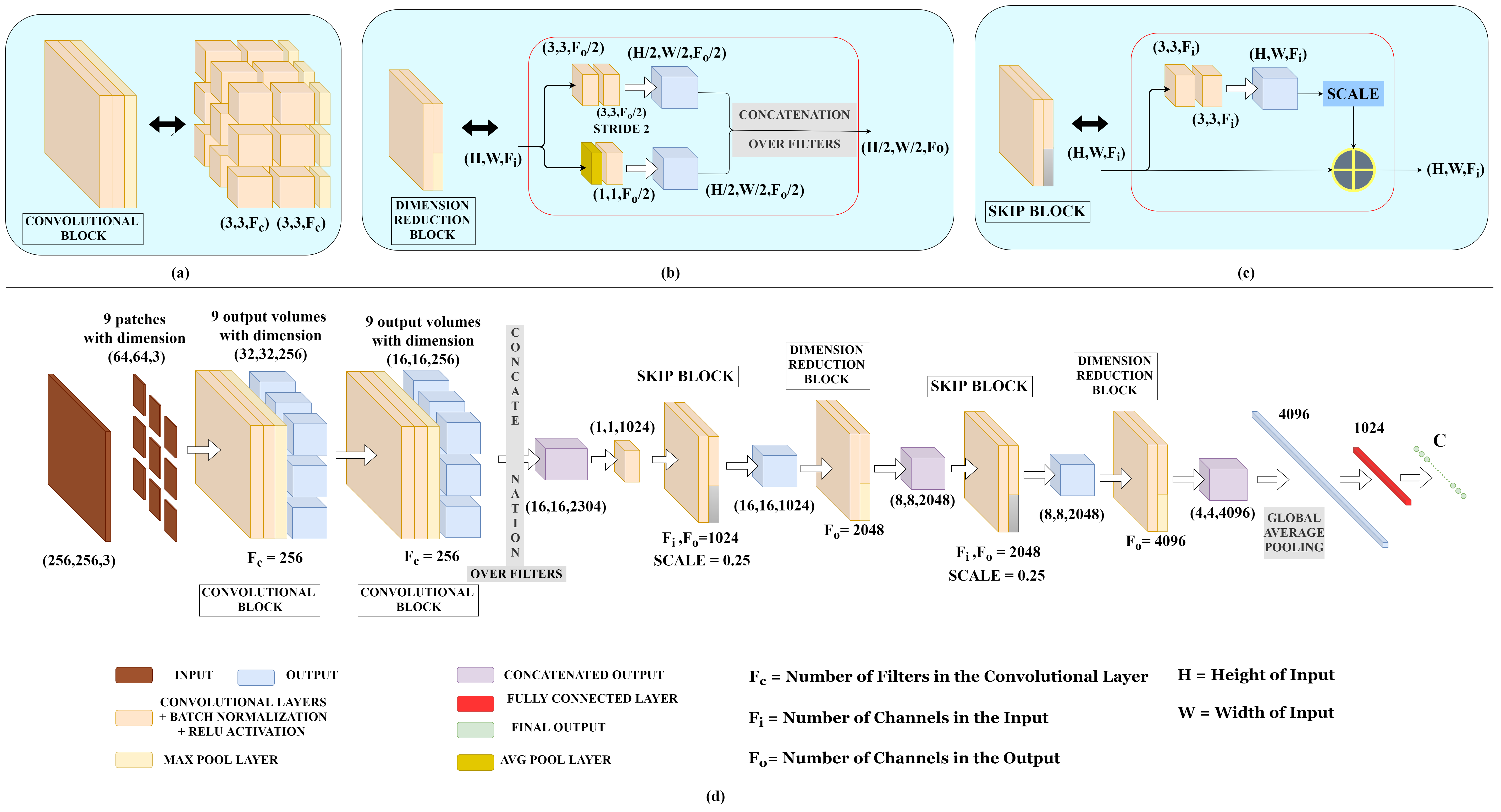}
    \caption{Proposed model for pretext task.  The model architecture contains three types of blocks: (a) Convolutional, (b) Dimension Reduction and (c) Skip. These three blocks are presented in expanded view above the model diagram and marked as (a), (b) and (c). The pretext model architecture is shown in (d). The image is best visible in 200\% scale.}
    \label{fig:pretext}
\end{minipage}
\end{figure*}

\indent In our paper, we have proposed a semi-parallel architecture (SKID) for the pretext task of jigsaw puzzle solving. The architecture is described as semi-parallel because of the nature of its design. The initial section of the architecture consists of 9 parallel branches of convolutional layers. This parallel nature of the convolutional branches does not continue to the end. The 9 parallel convolutional branches merge mid-way and the output features are processed by only a single branch of layers. The final network model (SKID-v3) architecture is used for the pretext task in our experiments as shown in Fig. \ref{fig:pretext}. As we have divided the input image into 9 patches, we feed each of the 9 patches into one of the 9 parallel convolutional branches. Each convolutional branch consists of two convolutional blocks. Each convolutional block again consists of two convolutional layers followed by a maxpooling layer. Each of the two convolutional layers has 256 filters of dimensions $3 \times 3$. The outputs obtained from the 9 convolutional branches are concatenated to form a resultant output of shape $16 \times 16 \times 2304$. This output is fed to a $1 \times 1$ convolutional layer with 1024 filters to reduce the dimensionality of the output. There are two main types of blocks in our architecture, \textit{Skip} block and \textit{Dimension Reduction} block. The architectures of both the blocks are shown in Fig. \ref{fig:pretext}. The output from the \textit{Skip} block is the summation of its input and the scaled output from two successive convolutional layers with filters of dimension $3 \times 3$. The scale factor in the \textit{Skip} block influences the performance of the model. In the \textit{Dimension Reduction} block, we use a combination of strided convolutions and average pooling to reduce the spatial dimension. The \textit{Dimension Reduction} block consists of two branches, as shown in Fig. \ref{fig:pretext}. The upper branch consists of two convolutional layers, with the first and second layer having strides 1 and 2, respectively. The lower branch consists of an average pooling layer followed by a $1 \times 1$ convolutional layer. The number of filters in all the convolutional layer in \textit{Dimension Reduction} block is same as the number of channels in the input. The outputs from the two branches are concatenated to obtain the final output with reduced spatial dimensions. The architecture of our \textit{Pretext} model consists of two of each such blocks, with a \textit{Dimension Reduction} block succeeding every \textit{Skip} block, as shown in the Fig. \ref{fig:pretext}. The output of the second \textit{Dimension Reduction} block is of the shape $4 \times 4 \times 4096$. Global Average Pooling is applied to this output to obtain an output of dimension $4096$, which is then fed to a  fully connected layer with 1024 nodes. Finally, another fully connected layer with $\mathcal{C}$ nodes is connected to the last layer, where $\mathcal{C}$ is the number of classes in the classification task.

\indent The Skip block is inspired from ResNet \cite{resnet} skip connections. In spatio-temporal data like MR data, the extent of occurrence of some desired feature is short both spatially and temporally. This requires the model to capture the intricate features in order to learn representations which can yield performance comparable to supervised learning algorithms in downstream tasks. As the extent of the desired features are short, the Skip blocks allows the model to learn the essential features by utilising the residual signals. The skip connections allow the model to prevent vanishing gradient problem and also utilise the features from initial layers in the higher layers for feature learning. The Dimension Reduction block aims at learning the downsampling procedure and thereby preventing information loss. This block also contains an average pooling layer, which is not learnable. Thus, this block combines a fixed operation with a learnable operation to capture more responsive features along with an increased expressibility. The Dimension Reduction block also serves as a bottleneck layer as the model is trained to learn the downsampling operation, which trains it to learn useful representations and discard redundant information. The effect of adding the Skip block and the Dimension Reduction block is evident from Table \ref{tab:super_comp_acc} and \ref{tab:contr_comp_auc}. in Sec. \ref{subsec:com_super}, where we present the performance of our model without the Skip and the Dimension Reduction block (Ours-NoBlocks). In the model named "Ours-NoBlocks", we replaced the Skip block with an identity function $f(x) = x$ and the Dimension Reduction Block was replaced by an average pooling layer with a window of dimensions $2 \times 2$.

\subsection{Downstream Task}
\label{subsec:downstream}
\indent The downstream task is an imbalanced multi-label classification task. We model the downstream task as 3 separate binary classification tasks for 3 classes, in accordance to the Binary Relevance method \cite{Godbole2004DiscriminativeMF}. We trained a different model for each of the three different planes and then used an ensemble of the models for the final prediction. 

\subsubsection{Ensembling Strategy}
\label{subsubsec:downstream_ensembling}
In this work, we use ensembling of the three models trained on three planes of MR data: Sagittal, Coronal and Axial, to obtain the final results. The main reason for doing so, instead of training a single model on all the three planes is that the nature of the images in the three planes are different. This causes the distribution of the images from the three planes to differ spatially. The details of the ensembling strategy used in the downstream task are given below.

\paragraph{Ensembling}
 For the ensemble, we used weighted majority voting \cite{ensemblebook} approach. It is essential to give more weight to the stronger classifier for each class. The output prediction of the ensemble of the 3 models is given by 
\begin{equation}
    H_j(x) = \sum_{i=1}^{T}w_i^j h_i^j(x) \geq 0.5,
\end{equation}
where $w_i^j$ is the weight assigned to the classifier $h_i^j$ for the $j^{th}$ class and T is the total number of classifiers. The weights are non-negative and are constrained by $\sum_{i=1}^{T} w_i^j = 1$. The weights are calculated as 
\begin{equation}
    w_i^j = log(\frac{p_i^j}{1-p_i^j}),
\end{equation}
where $p_i^j$ is the prediction accuracy for the $j^{th}$ class by the $i^{th}$ classifier.\\
\indent From the gradient activation class mappings obtained after training the downstream models on the three planes - Sagittal, Coronal and Axial (shown in Fig. \ref{fig:GradCAMDownstream}), we found out that the features contributing more towards the final predictions are distributed normally along the time axis. Based on this, for evaluation of the downstream model, we sampled the frames of the MR video from a normal distribution with $\mu = \frac{N_F}{2}$ and $\sigma = \frac{N_F}{4}$, where $N_F$ is the number of frames in the MR video. In addition to the sampling strategy, we also sampled 16 frames 8 times from each clip and obtained 8 predictions for each clip. We used the average of the 8 predictions as the final predictions for that clip.

\subsubsection{Model Architecture}
\label{subsubsec:downstream_model_arch}

\begin{figure*}[ht]
\begin{minipage}{\textwidth}
    \centering
    \includegraphics[width = 0.8\linewidth]{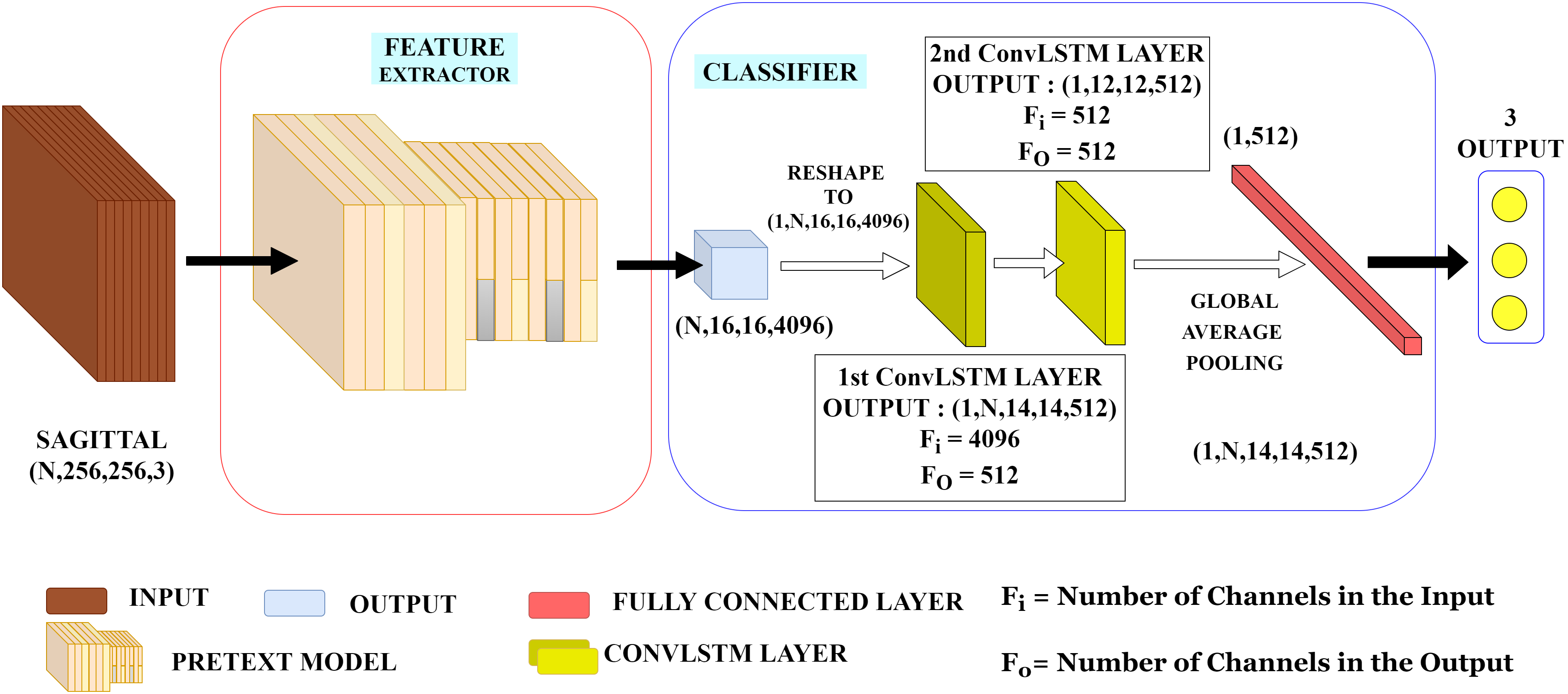}
    \caption{Network model used for the downstream task. The figure is best visible in 200\% scale.}
    \label{fig:downstream}
\end{minipage}
\end{figure*}

\indent The downstream model architecture (Fig. \ref{fig:downstream}) consists of the pretext model as the feature extractor and a Convolutional LSTM (ConvLSTM) network \cite{convlstm} as the classifier for efficient handling of the temporal correlation present in each clip. The downstream classifier consists of two ConvLSTM layers each with 512 output channels. The frames from the MR video clips are fed into the pretext model and the output obtained from the second \textit{Dimension Reduction} block is fed into the classifier. The 4D output tensor from the feature extractor is reshaped and passed to ConvLSTM network. Global Average Pooling is applied to the output obtained from the ConvLSTM network to get an output feature vector of 512 dimension. The final layer consists of 3 nodes for 3 classes with sigmoid activation function. The total number of parameters in the downstream model is approximately 103 Million.

\section{Experimental Details and Results}
\label{sec:exp}

\subsection{Dataset}
\label{subsec:dataset}

\begin{figure}
    \centering
    \includegraphics[scale=0.35]{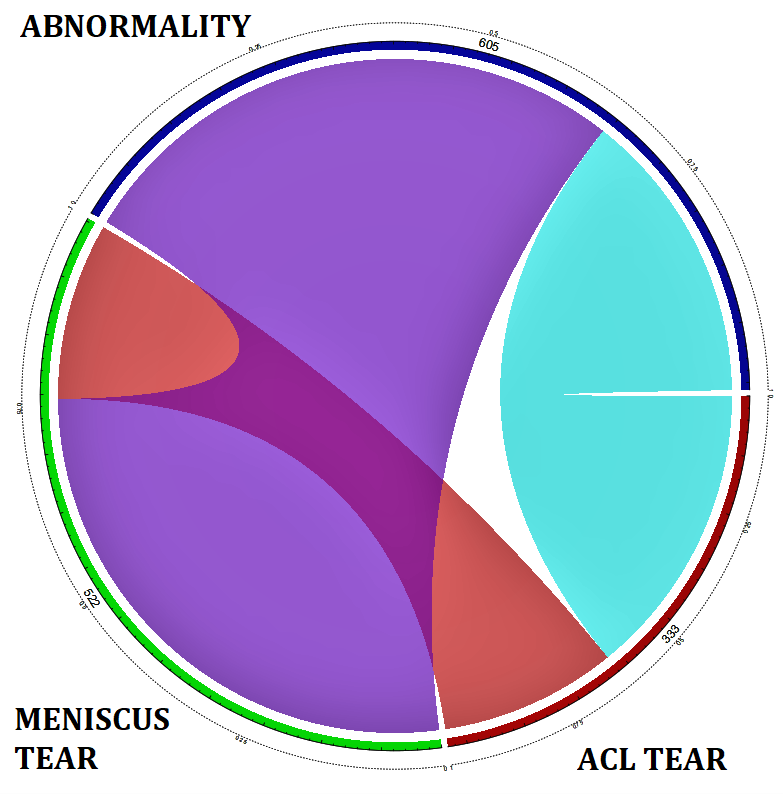}
    \caption{Concurrence plot of the labels in the MRNet dataset showing the interactions between different labels.}
    \label{fig:conplot}
\end{figure}

To validate the proposed self-supervised framework we use the benchmark MRNet \cite{mrnet} dataset, which contains 1370 clips. Out of the 1370 clips, 1130 clips are used as the training set and 120 clips are considered as tuning or validation set. The rest 120 clips are used for testing the model. The classes in the dataset are Abnormality, ACL Tear and Meniscus Tear. Out of the 1,130 training samples, 917 are Abnormal exams, 208 are ACL tears and 397 samples are Meniscus tears. The MRNet \cite{mrnet} dataset is an imbalanced multi-label dataset. The multi-label and imbalanced nature of the dataset can be observed in the concurrence plot in Fig. \ref{fig:conplot}. This gives us the opportunity to explore the effects of self-supervised learning techniques on a class imbalanced multi-label classification task. The average number of frames in each clip is 30.4. 

\indent We also used the KneeMRI \cite{kneemri} dataset for external validation. The dataset consists of 917 12-bit grayscale volumes of either left or right knees. Each volume record was labelled according to the ligament condition: (1) healthy, (2) partially injured, or (3) completely ruptured. The dataset contains 706 healthy, 174 partially injured and 55 completely ruptured samples. We use the dataset in two types of tasks: (1) binary classification task with the healthy and the partially injured samples as the negative samples and the completely ruptured samples as the positive samples, as mentioned in \cite{mrnet}, (2) imbalanced multi-class classification setting. The training, validation and test set splits used in the experiments are the same as \cite{mrnet}. %The different folders in KneeMRI \cite{kneemri} dataset, used for the training, validation and test set splits are given below (here the folder names are mentioned as 'vol01-10'):}
% \begin{itemize}
%     \item train split : 'vol08', 'vol04', 'vol03', 'vol09', 'vol06', 'vol07'
%     \item valid split : 'vol10', 'vol05'
%     \item test split : 'vol01', 'vol02'
% \end{itemize}

\subsection{Experiment on Pretext Task}
\label{subsubsec:exp_pre}

\subsubsection{Experimental Details}
\label{subsubsec:pre_exp_det}
\indent During training, the model was trained with frames chosen randomly according to a uniform distribution from MR videos. We optimized the categorical crossentropy loss of the model using RMSProp optimizer with a learning rate of $10^{-4}$ and exponential decay of 0.95 every epoch. We used a batch size of 16 during both training and validation stages. However, during the validation stage, we did not use any augmentations to the input. The pretext model was trained on a NVIDIA Tesla T4 on Google Colab. % and also on a 8GB NVIDIA GTX 2080Ti GPU. 
The training was stopped when the validation accuracy plateaued.

\subsubsection{Experimental Results}
\label{subsubsec:pre_exp_res}
\indent In this section, we present the experimental results of the proposed pretext model (Fig. \ref{fig:pretext}) on the three different planes, namely, Sagittal, Coronal and Axial, present in the MRNet dataset. The results are shown in Table \ref{tab:exp_pretext}. 

\begin{table*}[!ht]
    \centering
    \bgroup
    \caption{Experimental Results on the pretext task on different planes of Magnetic Resonance clips}
    \def\arraystretch{1.5}
    \begin{tabular}{c|c|c|c|c|c|c|c|c}
    \hline
       \multirow{2}{*}{Plane} & \multirow{2}{1.5cm}{\centering No. of parameters} & \multirow{2}{1cm}{\centering No. of classes} & \multirow{2}{2cm}{\centering Learning rate decay} & \multirow{2}{*}{Scale} & \multirow{2}{1cm}{\centering Batch size} & \multirow{2}{1.5cm}{\centering AWGN ($\mu,\sigma^2$)} & \multirow{2}{1.0cm}{\centering Epochs trained} & \multirow{2}{1.5cm}{\centering Validation Accuracy}\\
       {} &{} &{} &{} &{} &{} &{} &{} &{}   \\\hline \hline
        %Sagittal & 217.17 M & 500 & 0.95 / 1 epoch & 0.25 & 16 & (0.0,0.01) & 25 & 94.27\%\\ \hline
         Sagittal & 217.68 M & 1000 & 0.95 / 1 epoch & 0.25 & 16  & (0.0,0.01) & 25 & 88.39\%\\ \hline
         %Coronal & 217.17 M & 500 & 0.95 / 1 epoch & 0.25 & 16  & (0.0,0.01) & 20 & 90.31\%\\ \hline
         Coronal & 217.68 M & 1000 & 0.95 / 1 epoch & 0.25 & 16  & (0.0,0.01) & 20 & 87.72\% \\\hline
         %Axial & 217.17 M & 500 & 0.95 / 1 epoch & 0.25 & 16  & (0.0,0.01) & 10 & 95.52\%\\ \hline
         Axial & 217.68 M & 1000 & 0.95 / 1 epoch & 0.25 & 16  & (0.0,0.01) & 10 & 92.29\%\\ \hline
    \end{tabular}
    \label{tab:exp_pretext}
    \egroup
\end{table*}

\subsection{Experiment on Downstream task}
\label{subsec:exp_downstream}

\subsubsection{Experimental Details}
\label{subsubsec:down_exp_det}
\indent In this paper, besides solving a multi-label classification task, we also aim to examine how useful are the representations learnt by the pretext model. % from solving the pretext task alone. 
To fulfill our objectives, the weights of the pretext model were frozen and only the classifier in the downstream model was trained.\\ %Fine-tuning the weights disturbs the equilibrium reached by the pretext model and also alters the representations learnt during the pretext training phase.\\
\indent The downstream model was trained on 16 GB NVIDIA P100 GPU. We also wish to understand the temporally correlated features in the MR video and how it affects the inference. Thus, we chose only 16 frames, which is almost half of the average frame length (30.4), randomly according to an uniform distribution from each clip for training the downstream model. \\
\indent The downstream model with only 103 million parameters was trained by optimizing a weighted binary cross-entropy loss with a very low learning rate starting from $10^{-5}$ and decayed by 0.95 every epoch for a maximum of 20 epochs. The augmentations applied on the chosen frames are the same as those applied in the pretext experiments. For each plane in the dataset, we trained a different model. An ensemble of the three models was done using the weighted majority voting approach as described in Section \ref{subsubsec:downstream_ensembling} to obtain the final predictions. 

\subsubsection{Experimental Results}
\label{subsubsec:down_exp_res}

\begin{figure*}[h]
    \centering
        \begin{tabular}{cccc}
        \includegraphics[width = 0.285in]{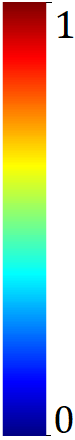}&
        
           \subfloat[Abnormality]{\includegraphics[width = 1.7in]{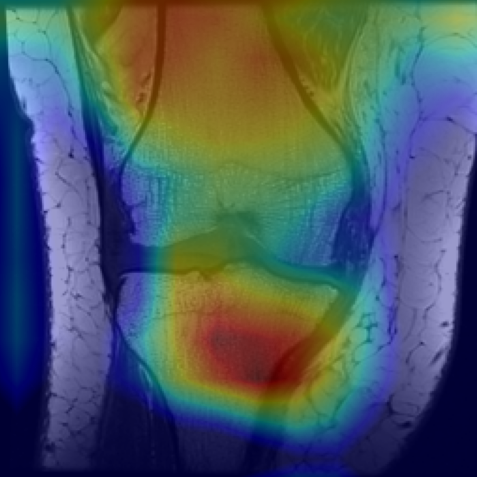}} &
           
            \subfloat[Abnormality]{\includegraphics[width = 1.7in]{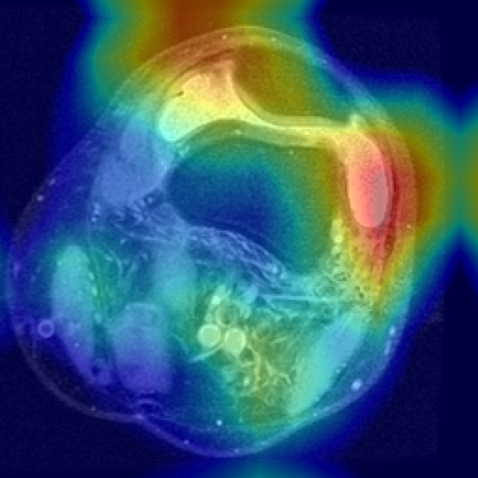}}&
            
        \subfloat[ACL Tear]{\includegraphics[width = 1.7in]{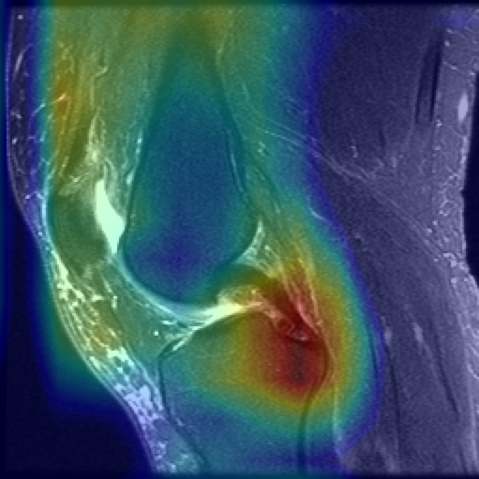}} \\
        
           \includegraphics[width = 0.285in]{jet_colormap_colorbar_2.png}&
           
           \subfloat[ACL Tear]{\includegraphics[width = 1.7in]{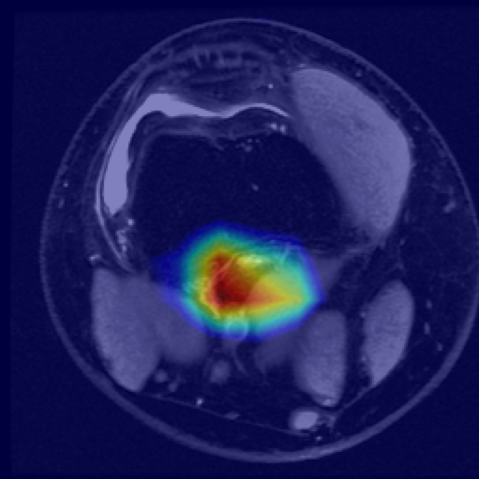}}&
            
        \subfloat[Meniscus Tear]{\includegraphics[width = 1.7in]{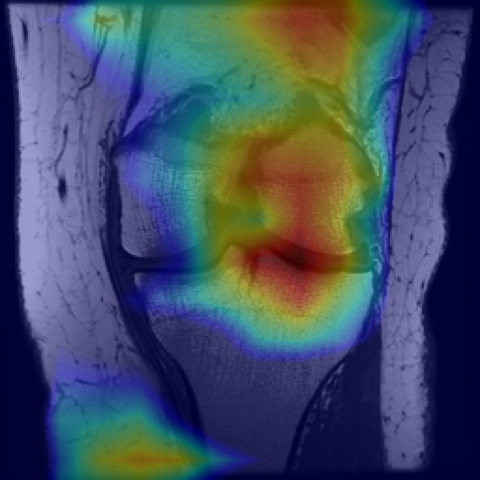}} &
        
            \subfloat[Meniscus Tear]{\includegraphics[width = 1.7in]{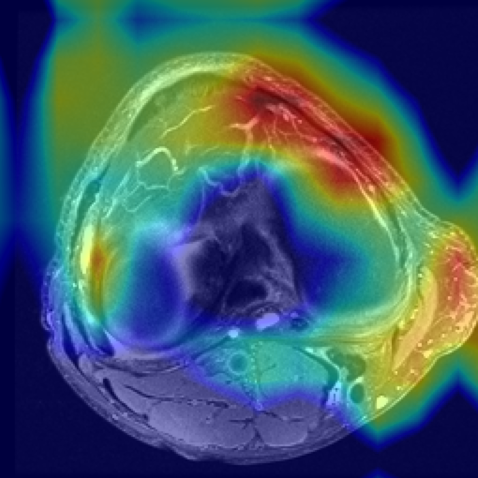}}\\
        \end{tabular}
    \caption{Gradient class activation mappings showing salient regions for different conditions in the Knee MR videos, obtained from the last Dimension Reduction block in the frozen pretext model of the downstream task.}
    \label{fig:GradCAMDownstream}
\end{figure*}

In this subsection, we present the results of our experiments on the 3 planes - Sagittal, Coronal and Axial of the MRNet dataset. We use both binary accuracy and AUC score as metrics for our model. In Table \ref{tab:evaluation}, we present the results of the three downstream models and the ensemble of those models for each of the three classes- Abnormality, ACL Tear and Meniscus Tear on the validation set of the MRNet dataset. During evaluation on the validation set we chose the frames according to the sampling strategy described in Section \ref{subsubsec:downstream_ensembling}. To analyse the efficiency and reliability of our method, we show the gradient class activation mappings \cite{gradcam} for the detection of all the three classes in Fig. \ref{fig:GradCAMDownstream}. The salient regions in Fig. \ref{fig:GradCAMDownstream} are the regions where the pretext model gains maximum information, which are then fed to the ConvLSTM network in the downstream task.\\
\indent
The weights learnt in the pretext task serve as a good initiation point for the downstream model and result in good convergence if compared to supervised or other SOTA contrastive algorithms as shown in Sec. \ref{subsec:com_super} and \ref{subsec:com_contr_algo}. The results show that even though the distribution of the features was imbalanced during the pretext task, the downstream result does not show any bias towards any particular label. This is evident from the fact that we only trained the ConvLSTM part in the downstream task and froze the parameters of the feature extractor, which was trained on the imbalanced multi-label dataset MRNet without any measures for dealing with the imbalance. As the number of positive occurrences for abnormality exceeds that of the other two labels, the model is expected to learn features mostly from the majority label. Consequently, this should affect the performance on the downstream task. However, we can observe in Tables \ref{tab:super_comp_auc} and \ref{tab:super_comp_acc} in Sec. \ref{subsec:com_super} that the performance of our model is at par with the supervised model.

\begin{table*}[!h]
    \centering
    \bgroup
    \caption{Evaluation results on validation set of MRNet dataset}
    \def\arraystretch{1.5}
    \begin{tabular}{c|c|c|c|c|c}
    \hline
        \multirow{2}{*}{Class} & \multirow{2}{*}{Plane} & \multirow{2}{2.5cm}{\centering Accuracy (5\%-95\% CI)}& \multirow{2}{2.5cm}{\centering Sensitivity (5\%-95\% CI)} & \multirow{2}{2.5cm}{\centering Specificity (5\%-95\% CI)} & \multirow{2}{2.5cm}{\centering AUC (5\%-95\% CI)}\\
        {}& {} & {} & {} & {} & {}\\ 
        \hline \hline
        \multirow{4}{*}{Abnormality}&Sagittal&
        0.883(0.869-0.896) & 0.968(0.956-0.979) &
        0.555(0.500-0.606) & 0.901(0.883-0.918)\\ \cline{2-6}
        {}&Coronal&0.860(0.843-0.875)&0.957(0.944-0.969)&
        0.474(0.423-0.529)&0.847(0.819-0.873)\\ \cline{2-6}
        {}&Axial&0.843(0.829-0.856)&0.947(0.935-0.958)&
        0.439(0.375-0.500)&0.867(0.839-0.897)\\ \cline{2-6}
        {}&Ensemble&\textbf{0.874}(0.862-0.887)&0.979(0.971-0.986)&
        0.486(0.432-0.543)& \textbf{0.904}(0.880-0.916)\\ \hline
        \multirow{4}{*}{ACL Tear}&Sagittal&
        0.740(0.720-0.758) & 0.630(0.597-0.665) & 0.833(0.807-0.862) & 0.848(0.828-0.867)\\ \cline{2-6}
        {}&Coronal&0.715(0.695-0.738)&0.793(0.758-0.822)&
        0.649(0.613-0.682)&0.813(0.791-0.828)\\ \cline{2-6}
        {}&Axial&0.807(0.785-0.826)&0.721(0.687-0.754)&
        0.879(0.858-0.906)&0.862(0.845-0.881)\\ \cline{2-6}
        {}&Ensemble&\textbf{0.800}(0.778-0.818)&0.740(0.709-0.769)&
        0.849(0.825-0.867)&\textbf{0.893}(0.878-0.909)\\ \hline
        \multirow{4}{*}{Meniscus Tear}&Sagittal&
        0.653(0.630-0.675) & 0.731(0.697-0.764) & 0.587(0.555-0.620) & 0.740(0.715-0.764)\\ \cline{2-6}
        {}&Coronal&0.717(0.698-0.736)&0.981(0.972-0.993)&
        0.517(0.484-0.551)&0.803(0.781-0.825)\\ \cline{2-6}
        {}&Axial&0.668(0.649-0.689)&0.748(0.713-0.785)&
        0.602(0.564-0.634)&0.744(0.718-0.768)\\ \cline{2-6}
        {}&Ensemble&\textbf{0.725}(0.706-0.746)&0.923(0.903-0.942)&
        0.574(0.539-0.608)&\textbf{0.810}(0.784-0.826)\\ \hline
         
    \end{tabular}
    \label{tab:evaluation}
    \egroup
    
\end{table*}

\subsection{External Validation}
\label{subsec:ext_valid}

\indent For both types of task for external validation, as mentioned in Sec. \ref{subsec:dataset}, we tested the capability of the proposed model to learn transferable features by comparing the performance in the downstream task, with the pretext encoder trained on the MRNet \cite{mrnet} and the KneeMRI \cite{kneemri} datasets. In the binary classification task setting, the downstream model trained with the pretext encoder that was trained on MRNet \cite{mrnet} dataset achieved an accuracy of 92.6\% and an AUC score of 0.761 on the test split of the KneeMRI \cite{kneemri} dataset. When the downstream model was trained with the pretext encoder that was trained on KneeMRI \cite{kneemri} dataset, the downstream model achieved an accuracy of 90.4\% and an AUC score of 0.741. The difference in performance can be attributed to the difference in the number of samples in the two datasets. As the number of samples in the MRNet \cite{mrnet} dataset is more than that in the KneeMRI \cite{kneemri} dataset, the pretext encoder learnt better features, which is reflected in the downstream performance. To deal with the imbalance in the number of samples for each binary label, we resorted to oversampling the minority class. In the imbalanced multi-class classification task setting, the downstream model trained with the pretext encoder that was trained on MRNet \cite{mrnet} and KneeMRI \cite{kneemri} datasets achieved an accuracy of 70.7\% and 68.6\% on the test split of the  dataset, respectively.

\subsection{Comparison with Supervised Algorithms}
\label{subsec:com_super}

\indent To compare our algorithm with supervised algorithms we replicated the MRNet \cite{mrnet} model and trained only on 16 frames chosen randomly from the clips according to an uniform distribution, like in our downstream experiments. The MRNet model was evaluated on frames chosen using the same sampling strategy mentioned in Section \ref{subsubsec:downstream_ensembling}. We present the comparison results in Table \ref{tab:super_comp_auc} and Table \ref{tab:super_comp_acc}.

We can see that using only 16 frames which almost half of the average number of frames per clip, the result of our proposed model SKID is comparable with the original MRNet \cite{mrnet} model. %Also if the AlexNet \cite{AlexNet} part of the MRNet (MRNet*) is frozen like the pretext model in our downstream task, the performance of MRNet drops significantly.
%and our model performs better in that scenario.

\begin{table}[ht]
    \centering
    \bgroup
    \caption{AUC Score Comparison with MRNet. }
    \def\arraystretch{1.5}
    \begin{tabular}{c|c|c|c|c}
    \hline
        \multirow{2}{1.2cm}{\centering Method} & \multicolumn{4}{c}{\centering AUC scores}\\ \cline{2-5}
        {} & Abnormality & ACL Tear & Meniscus Tear & Average\\  
        \hline \hline
        MRNet \cite{mrnet} & 0.944 & 0.915 & 0.822 & 0.894\\ \hline
        %MRNet* & 0.710 & 0.558 & 0.688 & 0.652\\ \hline
        Ours & \textbf{0.904} & \textbf{0.893} & \textbf{0.810} & \textbf{0.869}\\ \hline
        Ours-NoBlocks & 0.883 & 0.815 & 0.729 & 0.809 \\ \hline
         
    \end{tabular}
    \label{tab:super_comp_auc}
    \egroup
\end{table}

\begin{table}[ht]
    \centering
    \bgroup
    \caption{Accuracy Comparison with MRNet. }
    \def\arraystretch{1.5}
    \begin{tabular}{c|c|c|c|c}
    \hline
        \multirow{2}{1.2cm}{\centering Method} & \multicolumn{4}{c}{\centering Accuracy}\\ \cline{2-5}
        {} & Abnormality & ACL Tear & Meniscus Tear & Average\\  
        \hline \hline
        MRNet \cite{mrnet} & 0.850 & 0.867 & 0.725 & 0.814\\ \hline
        %MRNet* & 0.710 & 0.558 & 0.688 & 0.652\\ \hline
        Ours & \textbf{0.874} & \textbf{0.800} & \textbf{0.725} & \textbf{0.7997}\\ \hline
        Ours-NoBlocks & 0.825 & 0.733 & 0.608 & 0.722\\ \hline
         
    \end{tabular}
    \label{tab:super_comp_acc}
    
    \egroup
\end{table}

\subsection{Comparison with Contrastive Learning Algorithms}
\label{subsec:com_contr_algo}

In this subsection, we attempt to compare our novel task-specific architecture to some of the  state-of-the-art contrastive learning algorithms. One important point worthy to mention before proceeding further is that the experiments conducted using the contrastive learning algorithms were constrained by the availability of accelerator memory. The encoder in SimCLR \cite{simclr} or MoCo-v2 \cite{mocov2} has similar structure as used in the experiments in PIRL \cite{pirl} to account for the jigsaw transformation. Thus, the implementations of SimCLR \cite{simclr} and MoCo-v2 \cite{mocov2} deviates only in terms of the encoder structure in our experiments. The originality of the PIRL \cite{pirl} algorithm and encoder structure was preserved to the fullest. However, the size of the memory bank had to be reduced to 1024 because of memory constraint. Apart from these, no other experimental configuration was altered for the training of the encoders in the different contrastive learning algorithms. The primary transformation in the pretext experiments was \textit{Jigsaw} transformation, similar to PIRL \cite{pirl}.

\begin{table}[!ht]
    \centering
    %\begin{tabular}{cc}
    %\begin{minipage}{0.45\linewidth}
    \centering
    \bgroup
    \caption{AUC Score Comparison with Contrastive Learning algorithms. }
    \label{tab:contr_comp_auc}
    \def\arraystretch{1.5}
    \begin{tabular}{c|c|c|c|c}
    \hline
        \multirow{2}{1.2cm}{\centering Method} & \multicolumn{4}{c}{\centering AUC Scores}\\ \cline{2-5}
        {} & Abnormality & ACL Tear & Meniscus Tear & Average\\  
        \hline \hline
        PIRL \cite{pirl} & 0.653 & 0.655 & 0.590 & 0.633\\ \hline
        SimCLR \cite{simclr} & 0.676 & 0.691 & 0.663 & 0.677\\ \hline
        MocoV2 \cite{mocov2} & 0.682 & 0.389 & 0.600 & 0.557\\ \hline
        %MRNet* & 0.710 & 0.558 & 0.688 & 0.652\\ \hline
        Ours & \textbf{0.904} & \textbf{0.893} & \textbf{0.810} & \textbf{0.869}\\\hline
         
    \end{tabular}
    \egroup
    %\end{minipage}
    
    %\qquad
\end{table}
\quad
\begin{table}[!ht]
    %\begin{minipage}{0.45\linewidth}
    \centering
    \bgroup
    \caption{Accuracy Score Comparison with Contrastive Learning algorithms. }
    \label{tab:contr_comp_acc}
    \def\arraystretch{1.5}
    \begin{tabular}{c|c|c|c|c}
    \hline
        \multirow{2}{1.2cm}{\centering Method} & \multicolumn{4}{c}{\centering Accuracy}\\ \cline{2-5}
        {} & Abnormality & ACL Tear & Meniscus Tear & Average\\  
        \hline \hline
        PIRL \cite{pirl} & 0.750 & 0.600 & 0.567 & 0.639\\ \hline
        SimCLR \cite{simclr} & 0.742 & 0.625 & 0.592 & 0.653\\ \hline
        MocoV2 \cite{mocov2} & 0.733 & 0.458 & 0.558 & 0.583\\ \hline
        %MRNet* & 0.710 & 0.558 & 0.688 & 0.652\\ \hline
        Ours & \textbf{0.874} & \textbf{0.800} & \textbf{0.725} & \textbf{0.7997}\\ \hline
         
    \end{tabular}
    
    \egroup
    %\end{minipage}
    
    %\end{tabular}
\end{table}

\indent
Furthermore, the downstream experiments were conducted with the same experimental configuration as our model, i.e., we used a ConvLSTM network to account for the temporal dimension in the data which took as input, the output from the last convolutional layer of the pretext model, discarding the projector MLP, if any. Also, the parameters in the pretext model was frozen. The above steps were taken to ensure a fair comparison. The focus has been given mainly on the jigsaw transformation as the novelty of our work lies in utilising this transformation to build a novel architecture for learning representations from the data and showing how our method performs better than the state-of-the-art algorithms when using this transformation.\\
\indent
The AUC and accuracy scores of the different state-of-the-art contrastive learning algorithm implementations as mentioned above, for the downstream task of multi-label classification of the MR videos into the labels - Abnormality, ACL Tear and Meniscus Tear, are presented in Table \ref{tab:contr_comp_auc} and \ref{tab:contr_comp_acc}, respectively.

\section{Shortcomings of geometric transformation prediction task}
\label{sec:effects_GTP_pre}

\begin{figure*}[ht]
    \centering
    \begin{tabular}{cccc}
            \includegraphics[width = 0.285in]{jet_colormap_colorbar_2.png}&
            \subfloat[Translation]{\includegraphics[width = 1.7in]{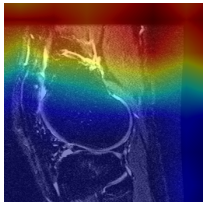}} &
            %\subfloat[Both Translation and Rotation]{\includegraphics[width = 0.85in]{gtp heatmap sagittal 27 classes _2.png}} &
            \subfloat[Rotation]{\includegraphics[width = 1.7in]{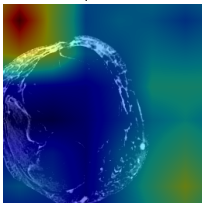}}&
            \subfloat[Translation]{\includegraphics[width = 1.7in]{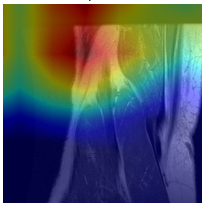}} \\
            
            \includegraphics[width = 0.285in]{jet_colormap_colorbar_2.png}&
            \subfloat[Both Translation and Rotation]{\includegraphics[width = 1.7in]{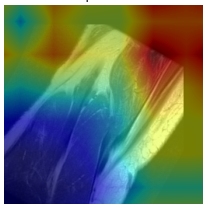}} &
            %\subfloat[Rotation and Scaling]{\includegraphics[width = 0.85in]{gtp heatmap coronal 27 classes_3.png}}\\
            %\includegraphics[width = 0.135in]{jet colormap colorbar 2.png}&
            \subfloat[Translation]{\includegraphics[width = 1.7in]{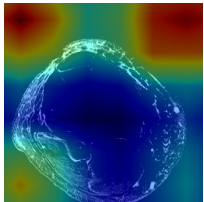}} &
            
            \subfloat[Both Rotation and Translation]{\includegraphics[width = 1.7in]{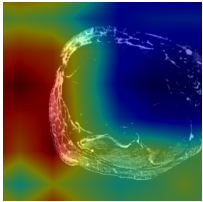}} \\
            %\subfloat[Both Rotation and Translation]{\includegraphics[width = 0.85in]{gtp heatmap axial for 27 classes_3.png}}\\
    \end{tabular}
    \caption{Gradient class activation mappings of frames from 3 planes, showing the regions of interest in a Geometric Transformation Prediction task (2 instances from each plane are shown). (a) and (b), (c) and (d), and (e) and (f) belong to Sagittal, Coronal and Axial planes, respectively. The individual captions indicate the geometrical transformation applied to the frames.}
    \label{fig:GTPGradCAM}
\end{figure*}

\indent Visual biomedical data like ultrasound videos, magnetic resonance videos, X-Ray images are essentially grayscale images. The features of interest are only a few pixels wide. Besides, these features are also present only for a few frames in spatio-temporal biomedical data. Thus, to learn these features from biomedical data, it requires very intricate and fine-grained learning. In addition to the temporal prevalence of the features, the spatial distribution of the pixels play an important role in representation learning. As the frames in videos of biomedical data are grayscale, the boundary transition regions from high intensity pixels to low intensity pixels pose problems in feature learning.\\
\indent In self-supervised learning methods like rotation prediction or geometric transformation prediction, the boundary pixels are relocated to new position leaving voids in their original locations. Because of this, the boundary pixels having higher intensity give rise to formation of some low level features. These features, e.g. blank areas, image boundaries, image corner points, etc. in the transformed MR video frames help the pretext model in optimizing the objective function without learning useful representations. Although geometric transformation prediction works well for natural images \cite{rotnet,videorotnet}, the behaviour was not the same in our experiments with MR data.\\
\indent To support the above findings on Geometric Transformation Prediction, we modeled a classification task and chose the transformation from a finite set $\mathcal{G}$, which can be expressed as a Cartesian product of four finite sets $\mathcal{R},\mathcal{T}_x,\mathcal{T}_y$ and $\mathcal{S}$, i.e. $\mathcal{G} = \mathcal{R} \times \mathcal{T}_x \times \mathcal{T}_y \times \mathcal{S}$ where $\mathcal{R} =\{-15^{\circ}, 0, 15^{\circ}\}$, $\mathcal{T}_x = \{-\lfloor0.1 L \rfloor,0,\lfloor0.1 L \rfloor\}$, $\mathcal{T}_y = \{-\lfloor0.1 L \rfloor,0,\lfloor0.1 L \rfloor\}$ and $\mathcal{S} = \{1,1.2\}$. Here $\mathcal{R},\mathcal{T}_x,\mathcal{T}_y$ and $\mathcal{S}$ denote the finite sets of angles of rotation in degrees, magnitude of translation along x-axis and y-axis in pixels and scale factors, respectively. $L$ denotes the dimension of each side of the frames. Total number of combinations of geometric transformations is 54. The pretext model used for testing the credibility of the geometric transformation prediction task, consisted of a pre-trained VGG Net with 54 outputs as we have used a combination of 54 transformations in total. The images in Fig. \ref{fig:GTPGradCAM} show the gradient class activation mappings \cite{gradcam} on a few frames from each plane (Sagittal, Coronal and Axial) after a random transformation is applied. We can clearly see that the salient regions are mainly concentrated on void regions, boundary pixels or corners. Choosing the proper pretext task is the most important part and the results shown in Fig. \ref{fig:GTPGradCAM} have greatly influenced in shaping our work.

% \section{\textcolor{red}{\st{Explainability in Self-supervised Learning}}}
% \label{sec:explainability}

%\textcolor{red}{\st{Deep learning models are approximations of some complex mapping functions from the input to the output space. Generally, deep learning models use hierarchical non-linear learning functions on the input data to obtain the desired output. However, the learning process and the convergence point is dependent on the random initialization point of the training process as well as the hyper-parameters. In other words, a deep learning model lacks explainability in itself. However, to further improve the performance, we need to understand the feature learning process of the deep learning models. Understanding what the deep learning models see at each layer and what it outputs, can give us some insight into the learning process. One such technique is gradient class activation mapping}} \cite{gradcam} \textcolor{red}{\st{which shows the magnitude of response obtained from different regions of the input.}}\\

\indent In self-supervised learning, no ground truth is used for learning representations. In absence of ground truth data, the proposed self-supervised framework makes use of pseudo-labels created for the pretext task. The performance in the downstream task is a measure of the quality of the representations learnt in the pretext task. The learning process and the decision taken by the downstream model in the inference stage depend on the features extracted by the pretext model, as we already mentioned in Sec. \ref{subsubsec:down_exp_det} that the pretext model was kept frozen in the downstream task. Thus, the gradient class activation mappings obtained from the last Dimension Reduction block of the pretext model is a direct reflection of the features extracted by the pretext model. In the gradient class activation mappings in Fig. \ref{fig:GradCAMDownstream}, we can see that the regions of interest important taking the correct decision in inference. In Sec. \ref{sec:effects_GTP_pre}, we can observe from the Fig. \ref{fig:GTPGradCAM}, that the regions of interest are not aligned with any features of interest in the MR frames, which explains why the standalone geometric transformation prediction task failed to learn any meaningful representations. It can be observed that the regions of interest in the GradCAM \cite{gradcam} outputs are mostly edges, empty regions or corners, which are low level features. Thus the model utilises the boundary discontinuities to classify the geometric transformations in the grayscale medical images. In other words, the model learns a shortcut path to obtain a global minimum in the loss landscape.

\section{Ablation Studies}
\label{sec: more_abl}

\subsection{Ablation on Pretext Model Architecture}
\label{subsec:ablation_studies}

\begin{table*}[!h]
    \centering
    \bgroup
    \caption{Different Variants of the Pretext model}
    \def\arraystretch{1.5}
    \begin{tabular}{c|c|c|c|c|c|c|c|c|c}
    \hline
       \multirow{4}{1.5cm}{\centering SKID Model variant} &
       \multicolumn{8}{c|}{Number of Filters in} &
       \multirow{4}{1.5cm}{\centering No. of parameters}\\ \cline{2-9}
       
       {} &
       \multirow{3}{*}{\parbox{1.1cm}{\centering Convo- lutional Block}} &
       \multirow{3}{0.8cm}{\centering $1 \times 1$ Conv. layer} &
       \multicolumn{2}{c|}{\centering $1^{st}$ Skip block} &
       \multirow{3}{1.4cm}{\centering $1^{st}$ Dim. Red. Block Output} &
       \multicolumn{2}{c|}{$2^{nd}$ Skip block} &
       \multirow{3}{1.4cm}{\centering $2^{nd}$ Dim. Red. Block Output} &
       {}\\ \cline{4-5} \cline{7-8}
       
       {} &
       {} &
       {} &
       \multirow{2}{1.4cm}{\centering $1^{st}$ Conv. layer} &
       \multirow{2}{0.9cm}{\centering Output} &
       {} &
       \multirow{2}{1.4cm}{\centering $1^{st}$ Conv. layer} &
       \multirow{2}{0.9cm}{\centering Output} &
       {} &
       {}\\
       
       {} & {} & {} &  {} & {} & {} & {} & {} & {} & {}\\\hline \hline
        
        SKID-v1 & 256 & 1024 & 512 & 1024 & 1024 & 512 & 1024 & 4096 & $\approx$ 109 M\\  \hline
        SKID-v2 & 256 & 1024 & 512 & 1024 & 2048 & 1024 & 2048 & 4096 & $\approx$ 170 M\\  \hline
        \multirow{2}{2cm}{\centering SKID-v3 (Proposed)} & \multirow{2}{*}{256} & \multirow{2}{*}{1024} & \multirow{2}{*}{1024} & \multirow{2}{*}{1024} & \multirow{2}{*}{2048} & \multirow{2}{*}{2048} & \multirow{2}{*}{2048} & \multirow{2}{*}{4096} & \multirow{2}{*}{$\approx$ 217 M}\\  
        {} & {} & {} &  {} & {} & {} & {} & {} & {} & {}\\\hline
    \end{tabular}
    \label{tab:pretext_ablation_filters}
    \egroup
   
\end{table*}

\begin{table*}[!ht]
    \centering
    \bgroup
    \caption{Ablation study on the pretext task for Sagittal plane of Magnetic Resonance videos. \dag The reported pretext validation accuracy is obtained at the epoch with the lowest validation loss.}
    \def\arraystretch{1.5}
    \begin{tabular}{c||c|c|c|c|c|c|c|c|c}
    \hline
        \multirow{2}{1.5cm}{\centering Indentifier}&
        \multirow{2}{1.5cm}{\centering SKID Model} & \multirow{2}{1.5cm}{\centering No. of parameters} & \multirow{2}{1.0cm}{\centering No. of classes} & \multirow{2}{1.5cm}{\centering Learning rate decay} & \multirow{2}{1.0cm}{Scale} & \multirow{2}{0.7cm}{\centering Batch size} & \multirow{2}{1.0cm}{\centering AWGN ($\mu,\sigma^2$)} & \multirow{2}{1.0cm}{\centering Epochs trained} & 
        \multirow{2}{1.5cm}{\centering Pretext Validation Accuracy \dag}\\
       {}&{} &{} &{} &{} &{} &{} &{} &{} &{}   \\\hline \hline
         %SKID-v1 & 108.8 M & 500 & 0.95/10 epochs & 1.0 & 4 & No & 200 & 86.46\%\\ \hline
        %  v1.1&SKID-v1 & 108.8 M & 500 & 0.95/1 epoch & 1.0 & 8  & -& 50 & 92.29\%\\ \hline
        %  v1.2&SKID-v1 & 108.8 M & 500 & 0.95/1 epoch & 0.25 & 8  & -&  100 & 92.6\%\\ \hline
         v1.1&\textcolor{black}{SKID-v1} & 108.8 M & 500 & 0.95/1 epoch & 0.25 & 16  & -&  50 & 90.67\%\\ \hline
         v1.2&\textcolor{black}{SKID-v1} & 108.8 M & 1000 & 0.95/1 epoch & 0.25 & 16  & -&  50 & 85.31\%\\ \hline \hline
         % SKID-v2 & 169.9 M & 500 & 0.95/1 epoch & 1.0 & 4  & No& 80 & 70.21\%\\ \hline
        %  v2.1&SKID-v2 & 169.9 M & 500 & 0.95/1 epoch & 1.0 & 16  & -& 20 & 90.16\%\\ \hline
         v2.1chang&\textcolor{black}{SKID-v2} & 169.9 M & 500 & 0.95/1 epoch & 0.25 & 16  & - & 50 & 89.53\%\\ \hline
         v2.2&\textcolor{black}{SKID-v2} & 169.9 M & 500 & 0.95/1 epoch & 0.25 & 16  & (0.0,0.01) & 50 & 88.90\%\\ \hline
         v2.3&\textcolor{black}{SKID-v2} & 170.4 M & \textcolor{black}{1000} & 0.95/1 epoch & 0.25 & 16  & - & 50 & 85.83\%\\ \hline
         v2.4&\textcolor{black}{SKID-v2} & 170.4 M & \textcolor{black}{1000} & 0.95/1 epoch & 0.25 & 16  & (0.0,0.01) & 50 & 85.57\%\\ \hline \hline
         % SKID-v3 & 217.17 M & 500 & 0.95/1 epoch & 0.25 & 16  & (0.0,0.05)& 21 & 88.13\%\\ \hline
         v3.1&\textcolor{black}{SKID-v3} & 217.17 M & 500 & 0.95/1 epoch & 0.25 & 16  & (0.0,0.01) & 50 & 94.27\% \\ \hline
         v3.2&\multirow{2}{2cm}{\centering \textcolor{black}{SKID-v3} (Proposed)} & \multirow{2}{*}{217.68 M} & \multirow{2}{*}{1000} & \multirow{2}{*}{0.95/1 epoch} & \multirow{2}{*}{0.25} & \multirow{2}{*}{16}  & \multirow{2}{*}{(0.0,0.01)} & \multirow{2}{*}{50} & \multirow{2}{*}{88.27\%} \\ 
         {}&{} & {}& {}& {}& {}& {}& {}& {}& {}\\
         \hline
    \end{tabular}
    \label{tab:ablation_sagittal}
    \egroup
    
\end{table*}

\indent This section aims to compare the performance of the pretext model with different configurations. We obtained these configurations by setting different values of the hyper-parameters like scale, learning rate decay schedule, batch size, etc. This subsection helps to understand how we reached our final model by repeated experimentation and tuning. SKID-v1 was used as the first model from which we started upgrading the model. In both, SKID-v1 and SKID-v2, a different variant of the \textit{Skip} block was used, where the number of filters in the first convolutional layer is half of the number of channels in its input. The number of output channels in \textit{Skip} and \textit{Dimension Reduction} blocks was also altered to get the different models. The details of the different variants of the architecture are given in the Table \ref{tab:pretext_ablation_filters}. The table shows the number of filters used in building the different architectures that we have experimented upon to obtain the final proposed model (SKID-v3).\\
\indent Table \ref{tab:ablation_sagittal} shows the results of the different models under different experimental setting on the pretext task. As we can see, the results improved as the number of parameters increased. A faster learning rate decay of 0.95 per epoch and a scale factor value of 0.25 proved to be optimal in our experiments. We also used Gaussian noise with mean 0.0 and variance 0.01 to make our model more robust. Moreover, increasing the batch size helped to achieve the convergence faster. The number of classes was increased from 500 to 1000 to train the pretext model to learn sparser features. In the downstream task, we used the model trained on 1000 classes as the learned representations are more generalized in nature. The justification for using the model trained with more number of classes in the pretext task has been justified by an ablation study in Sec. \ref{subsec:num_classes}. As the number of chosen arrangements increases, the possibility of any one convolutional branch to learn only some specific patches decreases. This is mainly because the patches become uniformly distributed over the 9 branches, according to the law of large numbers. %This has been discussed in furhter detail in Sec. \ref{subsec:num_classes}.

\begin{figure}
    \centering
    \includegraphics[width = \linewidth]{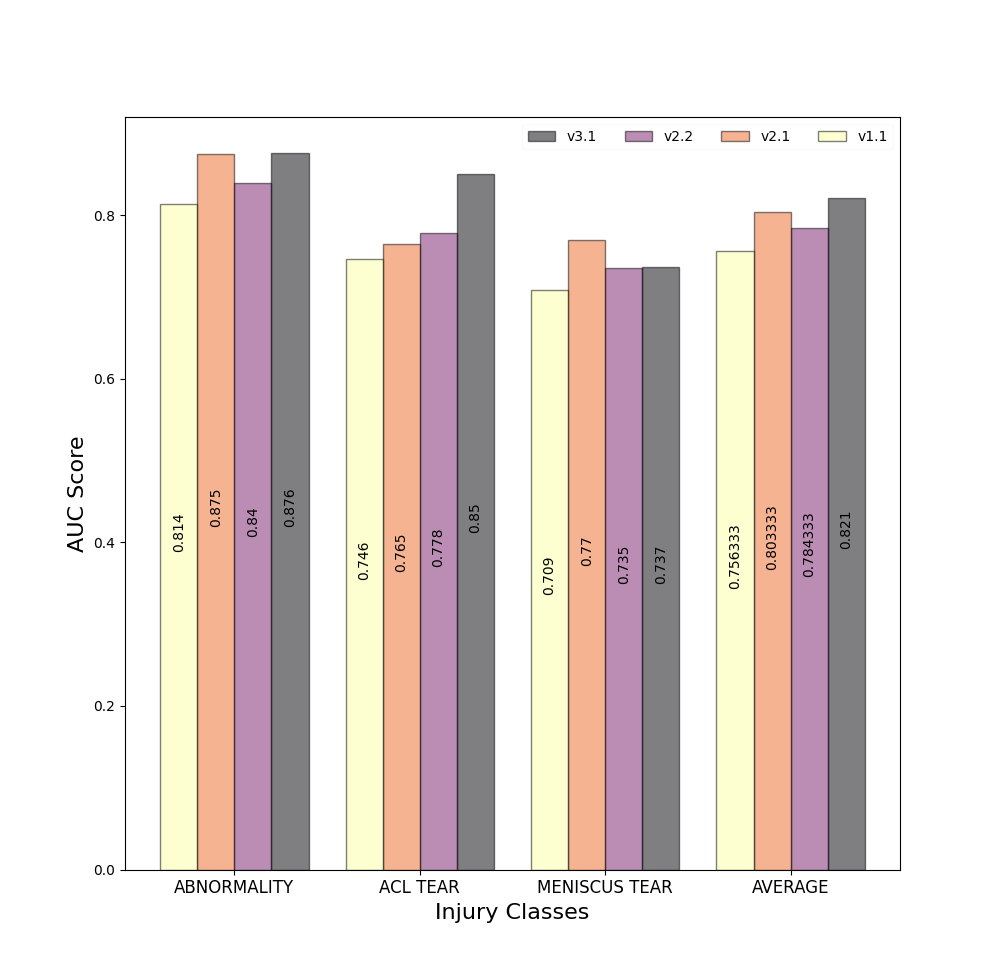}
    \caption{Comparison of AUC scores in Downstream task of Knee Injury classification using models trained with 500 classes in the pretext phase.}
    \label{fig:model_abl_arch_500}
\end{figure}

However, the primary goal in self-supervised learning is to ensure that the pre-trained models provides a better initialization point for the downstream task. To justify our choice of using SKID-v3 for the downstream experiments, we present the results of the SKID-v1, SKID-v2 and SKID-v3 models in the downstream task of Knee injury classification using only the Sagittal plane MR data. From the illustrative comparative results presented in Fig. \ref{fig:model_abl_arch_500}, we can see that the model SKID-v3 performed the best among all the three variations in terms of average AUC score. For this comparison, the models used in the comparison were trained for only 50 epochs and with a batch size of 16 in the pretext task to maintain consistency of experimental environment. We can infer that increasing the number of parameters of the self-supervised model improves performance on small scale medical datasets. As all the dataset classes are imbalanced as evident from Fig. \ref{fig:conplot}, we used the AUC scores instead of accuracy for comparison.

\subsection{Label Efficiency}
\label{subsec:label_eff}

\begin{figure*}[!h]
    \centering
    \includegraphics[width = \textwidth]{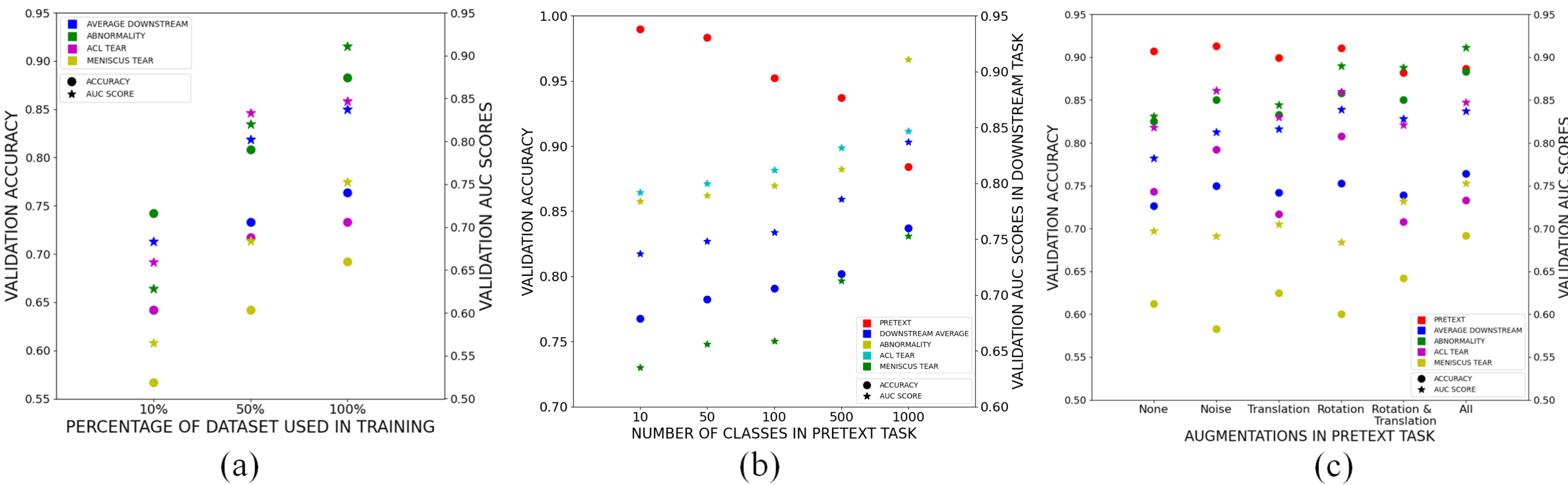}
    \caption{(a) Validation accuracy and AUC score of downstream models trained on 10\%, 50\% and 100\% of the dataset, (b) Effect of number of classes on validation accuracy of pretext and downstream tasks, (c) Effect of augmentations on pretext and downstream tasks. For (a), (b) \& (c), circle and asterisk symbols represent validation accuracy and AUC score, respectively. Different colors represent different cases.}
    \label{fig:more_ablation}
\end{figure*}

The quality of the representations learnt by the pretext model can be evaluated by observing its performance in semi-supervised tasks. As the dataset contains only 1130 samples, sampling 1\% data from it would only result in 11 samples in the training set. That is why we sample 10\% and 50\% of the training set to show the efficiency of our model. We follow the same experimental configuration as in the downstream task experiments, where we only train the classifier part and do not update the parameters in the feature extractor part.\\
\indent
In the Fig. \ref{fig:more_ablation}.a, both the validation accuracy and validation AUC scores are plotted for 10\%, 50\% and 100\% of the training data. It can be observed that, even with training on 10\% of the training data, our model performs better than random prediction. Calculating the performance metrics on random predictions, we found the AUC scores for the classes Abnormality, ACL Tear and Meniscus Tear to be 0.5416 (0.5008-0.5769 5\%-95\% CI), 0.5955 (0.5697-0.6213 5\%-95\% CI) and 0.4478 (0.4168-0.4808 5\%-95\% CI), respectively. Similarly, the accuracy scores for those classes are 0.5312 (0.5025-0.5556 5\%-95\% CI), 0.5752 (0.5509-0.5981 5\%-95\% CI), 0.4683 (0.4443-0.4945 5\%-95\% CI), respectively.
%Table -- shows the performance of other contrastive learning based self-supervised learning algorithms on the same fractions of training data as our model.

\subsection{Role of Chosen Number of Classes in the Pretext Task}
\label{subsec:num_classes}

The generalization capability of the pretext model is directly related to the number of jigsaw arrangements chosen during the training step. In Fig. \ref{fig:more_ablation}.b, we can observe that the AUC scores of the individual classes and also the average accuracy in the downstream task increases as the number of classes increases, even though the validation accuracy in the pretext task decreases. With the increase in the number of classes, the pretext task becomes more and more difficult, and hence the accuracy in the pretext task decreases. But it leads to better generalized representation learning and thereby results in better performance in the downstream task. 

\subsection{Effect of Augmentations}
\label{subsec:aug_eff}

In this subsection, we show the effect of different augmentations used in the Pretext task on the performance in the Downstream task. Fig. \ref{fig:more_ablation}.c shows the effects of the individual augmentations on the average downstream accuracy and AUC scores, as well as the performance on the individual classes. For the downstream experiments with the different pretext models trained with different augmentations, no augmentation was applied on the input samples. This helps in effectively demonstrating the effect of different augmentations on the representation learnt by the pretext model. The quality of the representations is ultimately reflected in the downstream task performance.

\subsection{ConvLSTM vs 3D CNN}
\label{subsec:convlstmvs3dcnn}

For video classification tasks, the use of 3D CNN has been effective. We also know that to learn features from temporal data, Recurrent Neural Networks like Long-Short Term Memory (LSTM) serve as an effective tool. Combining convolutional operation in LSTM, we get Convolutional LSTM (ConvLSTM) \cite{convlstm}. We followed the same experimental configuration for both the models containing ConvLSTM and 3D-CNN \cite{3dcnn}. The comparison results are provided in Table \ref{tab:comp_acc_convlstm_3dcnn} and \ref{tab:comp_auc_convlstm_3dcnn}. It is evident that the ConvLSTM performs better in out experiments and this is the main reason behind preferring ConvLSTM over 3D CNN.

\begin{table}[!h]
    \centering
    \bgroup
    \caption{Accuracy comparison between Downstream models with ConvLSTM and 3D CNN}
    \def\arraystretch{1.8}
    \begin{tabular}{c|c|c|c|c}
    \hline
        \multirow{2}{1.2cm}{\centering Method} & \multicolumn{4}{c}{\centering Accuracy}\\ \cline{2-5}
        {} & Abnormality & ACL Tear & Meniscus Tear & Average\\  
        \hline \hline
        %MRNet* & 0.710 & 0.558 & 0.688 & 0.652\\ \hline
        ConvLSTM & \textbf{0.874} & \textbf{0.800} & \textbf{0.725} & \textbf{0.7997}\\ \hline
        3D CNN & 0.842 & 0.767 & 0.708 & 0.7723\\ \hline
         
    \end{tabular}
    \label{tab:comp_acc_convlstm_3dcnn}
    \egroup
\end{table}

\begin{table}[!h]
    \centering
    \bgroup
    \caption{AUC score comparison between Downstream models using ConvLSTM and 3D CNN}
    \def\arraystretch{1.8}
    \begin{tabular}{c|c|c|c|c}
    \hline
        \multirow{2}{1.2cm}{\centering Method} & \multicolumn{4}{c}{\centering Accuracy}\\ \cline{2-5}
        {} & Abnormality & ACL Tear & Meniscus Tear & Average\\  
        \hline \hline
        %MRNet* & 0.710 & 0.558 & 0.688 & 0.652\\ \hline
        ConvLSTM & \textbf{0.904} & \textbf{0.893} & \textbf{0.810} & \textbf{0.869}\\ \hline
        3D CNN & 0.826 & 0.830 & 0.774 & 0.81 \\ \hline
         
    \end{tabular}
    \label{tab:comp_auc_convlstm_3dcnn}
    \egroup
\end{table}

\section{Conclusion}
\label{sec:conclusion}

\indent We proposed SKID, a self-supervised deep learning method for learning spatial context-invariant representations from MR video frames. The objective of this work is to explore the possibilities that self-supervised learning provides in medical image analysis domain. The challenges associated with the pretext task are discussed and analyzed thoroughly in our paper. The experimental results presented in this paper show that the quality of the representations that the pretext models learn depend on the self-supervised learning approaches. In the downstream task we successfully implemented a multi-label class imbalanced problem without fine-tuning the features learnt in the pretext task. This explains the quality of the representations learnt by the novel pretext architecture proposed in this paper. Different ablation studies are also included in the paper to get better insights of the work. To the best of our knowledge, this is the first work of its kind in showing the effectiveness and reliability of self-supervised learning algorithm in class imbalanced multi-label classification task on MR data.

{\small
\bibliographystyle{unsrt}
\bibliography{main}

\begin{thebibliography}{10}

\bibitem{alexnet}
A.~Krizhevsky, Ilya Sutskever, and Geoffrey~E. Hinton.
\newblock Imagenet classification with deep convolutional neural networks.
\newblock In {\em Communications of the ACM}, 2017.

\bibitem{vggnet}
Karen Simonyan and Andrew Zisserman.
\newblock Very deep convolutional networks for large-scale image recognition.
\newblock In {\em International Conference on Learning Representations}, 2015.

\bibitem{effnet}
Mingxing Tan and Quoc Le.
\newblock {E}fficient{N}et: Rethinking model scaling for convolutional neural
  networks.
\newblock volume~97 of {\em Proceedings of Machine Learning Research}, pages
  6105--6114, Long Beach, California, USA, 09--15 Jun 2019. PMLR.

\bibitem{sqandextnet}
Jie Hu, L.~Shen, Samuel Albanie, Gang Sun, and Enhua Wu.
\newblock Squeeze-and-excitation networks.
\newblock {\em IEEE Transactions on Pattern Analysis and Machine Intelligence},
  42:2011--2023, 2020.

\bibitem{densenet}
Gao Huang, Zhuang Liu, and Kilian~Q. Weinberger.
\newblock Densely connected convolutional networks.
\newblock {\em 2017 IEEE Conference on Computer Vision and Pattern Recognition
  (CVPR)}, pages 2261--2269, 2017.

\bibitem{resnet}
Kaiming He, X.~Zhang, Shaoqing Ren, and Jian Sun.
\newblock Deep residual learning for image recognition.
\newblock {\em 2016 IEEE Conference on Computer Vision and Pattern Recognition
  (CVPR)}, pages 770--778, 2016.

\bibitem{howtr}
Jason Yosinski, Jeff Clune, Yoshua Bengio, and Hod Lipson.
\newblock How transferable are features in deep neural networks?
\newblock In Zoubin Ghahramani, Max Welling, Corinna Cortes, Neil~D. Lawrence,
  and Kilian~Q. Weinberger, editors, {\em Advances in Neural Information
  Processing Systems 27: Annual Conference on Neural Information Processing
  Systems 2014, December 8-13 2014, Montreal, Quebec, Canada}, pages
  3320--3328, 2014.

\bibitem{surveytransl}
Fuzhen Zhuang, Zhiyuan Qi, Keyu Duan, Dongbo Xi, Yongchun Zhu, H.~Zhu, Hui
  Xiong, and Q.~He.
\newblock A comprehensive survey on transfer learning.
\newblock {\em ArXiv}, abs/1911.02685, 2019.

\bibitem{surveysemisup}
Jesper~E. van Engelen and H.~Hoos.
\newblock A survey on semi-supervised learning.
\newblock {\em Machine Learning}, 109:373--440, 2019.

\bibitem{surveysemisup2}
A.~Mey and M.~Loog.
\newblock Improvability through semi-supervised learning: A survey of
  theoretical results.
\newblock {\em ArXiv}, abs/1908.09574, 2019.

\bibitem{surveyweaksup}
J{\'e}r{\^o}me Rony, Soufiane Belharbi, J.~Dolz, I.~B. Ayed, L.~McCaffrey, and
  Eric Granger.
\newblock Deep weakly-supervised learning methods for classification and
  localization in histology images: a survey.
\newblock {\em ArXiv}, abs/1909.03354, 2019.

\bibitem{shuffleandlearn}
I.~Misra, C.~L. Zitnick, and M.~Hebert.
\newblock Shuffle and learn: Unsupervised learning using temporal order
  verification.
\newblock In {\em ECCV}, 2016.

\bibitem{temporder}
Fatemeh Siar, A.~Gheibi, and Ali Mohades.
\newblock Unsupervised learning of visual representations by solving shuffled
  long video-frames temporal order prediction.
\newblock {\em ACM SIGGRAPH 2020 Posters}, 2020.

\bibitem{oddoneout}
Basura Fernando, Hakan Bilen, E.~Gavves, and Stephen Gould.
\newblock Self-supervised video representation learning with odd-one-out
  networks.
\newblock {\em 2017 IEEE Conference on Computer Vision and Pattern Recognition
  (CVPR)}, pages 5729--5738, 2017.

\bibitem{cliporder}
D.~Xu, Jun Xiao, Zhou Zhao, J.~Shao, Di~Xie, and Y.~Zhuang.
\newblock Self-supervised spatiotemporal learning via video clip order
  prediction.
\newblock {\em 2019 IEEE/CVF Conference on Computer Vision and Pattern
  Recognition (CVPR)}, pages 10326--10335, 2019.

\bibitem{sssptemporder}
Himanshu Buckchash and Balasubramanian Raman.
\newblock Sustained self-supervised pretraining for temporal order
  verification.
\newblock In Bhabesh Deka, Pradipta Maji, Sushmita Mitra, Dhruba~Kumar
  Bhattacharyya, Prabin~Kumar Bora, and Sankar~Kumar Pal, editors, {\em Pattern
  Recognition and Machine Intelligence - 8th International Conference, PReMI
  2019, Tezpur, India, December 17-20, 2019, Proceedings, Part {I}}, volume
  11941 of {\em Lecture Notes in Computer Science}, pages 140--149. Springer,
  2019.

\bibitem{imgcolor}
Richard Zhang, Phillip Isola, and Alexei~A. Efros.
\newblock Colorful image colorization.
\newblock In Bastian Leibe, Jiri Matas, Nicu Sebe, and Max Welling, editors,
  {\em Computer Vision -- ECCV 2016}, pages 649--666, Cham, 2016. Springer
  International Publishing.

\bibitem{rotnet}
Spyros Gidaris, Praveer Singh, and Nikos Komodakis.
\newblock Unsupervised representation learning by predicting image rotations.
\newblock In {\em 6th International Conference on Learning Representations,
  {ICLR} 2018, Vancouver, BC, Canada, April 30 - May 3, 2018, Conference Track
  Proceedings}. OpenReview.net, 2018.

\bibitem{videorotnet}
Longlong Jing, Xiaodong Yang, Jingen Liu, and Y.~Tian.
\newblock Self-supervised spatiotemporal feature learning via video rotation
  prediction.
\newblock {\em arXiv: Computer Vision and Pattern Recognition}, 2018.

\bibitem{contextpred}
C.~Doersch, A.~Gupta, and Alexei~A. Efros.
\newblock Unsupervised visual representation learning by context prediction.
\newblock {\em 2015 IEEE International Conference on Computer Vision (ICCV)},
  pages 1422--1430, 2015.

\bibitem{contextenc}
Deepak Pathak, Philipp Kr{\"a}henb{\"u}hl, Jeff Donahue, Trevor Darrell, and
  Alexei~A. Efros.
\newblock Context encoders: Feature learning by inpainting.
\newblock {\em 2016 IEEE Conference on Computer Vision and Pattern Recognition
  (CVPR)}, pages 2536--2544, 2016.

\bibitem{jnoble}
Jianbo Jiao, Richard Droste, L.~Drukker, A.~T. Papageorghiou, and J.~Noble.
\newblock Self-supervised representation learning for ultrasound video.
\newblock {\em 2020 IEEE 17th International Symposium on Biomedical Imaging
  (ISBI)}, pages 1847--1850, 2020.

\bibitem{mococxr}
Hari Sowrirajan, Jingbo Yang, Andrew~Y. Ng, and Pranav Rajpurkar.
\newblock Moco pretraining improves representation and transferability of chest
  x-ray models.
\newblock In Mattias~P. Heinrich, Qi~Dou, Marleen de~Bruijne, Jan Lellmann,
  Alexander Schlaefer, and Floris Ernst, editors, {\em Medical Imaging with
  Deep Learning, 7-9 July 2021, L{\"{u}}beck, Germany}, volume 143 of {\em
  Proceedings of Machine Learning Research}, pages 728--744. {PMLR}, 2021.

\bibitem{moco}
Kaiming He, Haoqi Fan, Yuxin Wu, Saining Xie, and Ross~B. Girshick.
\newblock Momentum contrast for unsupervised visual representation learning.
\newblock In {\em 2020 {IEEE/CVF} Conference on Computer Vision and Pattern
  Recognition, {CVPR} 2020, Seattle, WA, USA, June 13-19, 2020}, pages
  9726--9735. {IEEE}, 2020.

\bibitem{noroozi}
Mehdi Noroozi and Paolo Favaro.
\newblock Unsupervised learning of visual representations by solving jigsaw
  puzzles.
\newblock In Bastian Leibe, Jiri Matas, Nicu Sebe, and Max Welling, editors,
  {\em Computer Vision -- ECCV 2016}, pages 69--84, Cham, 2016. Springer
  International Publishing.

\bibitem{damagedjig}
Dahun Kim, Donghyeon Cho, Donggeun Yoo, and In-So Kweon.
\newblock Learning image representations by completing damaged jigsaw puzzles.
\newblock {\em 2018 IEEE Winter Conference on Applications of Computer Vision
  (WACV)}, pages 793--802, 2018.

\bibitem{videojig}
U.~Ahsan, R.~Madhok, and Irfan Essa.
\newblock Video jigsaw: Unsupervised learning of spatiotemporal context for
  video action recognition.
\newblock {\em 2019 IEEE Winter Conference on Applications of Computer Vision
  (WACV)}, pages 179--189, 2019.

\bibitem{iterreorg}
Chen Wei, Lingxi Xie, Xutong Ren, Yingda Xia, Chi Su, Jiaying Liu, Q.~Tian, and
  A.~Yuille.
\newblock Iterative reorganization with weak spatial constraints: Solving
  arbitrary jigsaw puzzles for unsupervised representation learning.
\newblock {\em 2019 IEEE/CVF Conference on Computer Vision and Pattern
  Recognition (CVPR)}, pages 1910--1919, 2019.

\bibitem{Godbole2004DiscriminativeMF}
Shantanu Godbole and Sunita Sarawagi.
\newblock Discriminative methods for multi-labeled classification.
\newblock volume vol. 3056, 08 2004.

\bibitem{ensemblebook}
Zhi-Hua Zhou.
\newblock {\em Ensemble Methods: Foundations and Algorithms}.
\newblock Chapman \& Hall/CRC, 1st edition, 2012.

\bibitem{convlstm}
Xingjian Shi, Zhourong Chen, Hao Wang, Dit-Yan Yeung, Wai-kin Wong, and
  Wang-chun Woo.
\newblock Convolutional lstm network: A machine learning approach for
  precipitation nowcasting.
\newblock In C.~Cortes, N.~Lawrence, D.~Lee, M.~Sugiyama, and R.~Garnett,
  editors, {\em Advances in Neural Information Processing Systems}, volume~28,
  pages 802--810. Curran Associates, Inc., 2015.

\bibitem{mrnet}
Nicholas Bien, Pranav Rajpurkar, R.~Ball, J.~Irvin, A.~Park, Erik Jones,
  Michael Bereket, B.~Patel, K.~Yeom, K.~Shpanskaya, S.~Halabi, E.~Zucker,
  G.~Fanton, D.~Amanatullah, C.~Beaulieu, Geoffrey~M. Riley, R.~Stewart,
  F.~Blankenberg, D.~Larson, Ricky Jones, C.~Langlotz, A.~Ng, and M.~Lungren.
\newblock Deep-learning-assisted diagnosis for knee magnetic resonance imaging:
  Development and retrospective validation of mrnet.
\newblock {\em PLoS Medicine}, 15, 2018.

\bibitem{kneemri}
Ivan Stajduhar, Mihaela Mamula, Damir Miletic, and G{\"{o}}zde~B. {\"{U}}nal.
\newblock Semi-automated detection of anterior cruciate ligament injury from
  {MRI}.
\newblock {\em Comput. Methods Programs Biomed.}, 140:151--164, 2017.

\bibitem{gradcam}
R.~R. Selvaraju, Michael Cogswell, Abhishek Das, Ramakrishna Vedantam,
  D.~Parikh, and Dhruv Batra.
\newblock Grad-cam: Visual explanations from deep networks via gradient-based
  localization.
\newblock {\em 2017 IEEE International Conference on Computer Vision (ICCV)},
  pages 618--626, 2017.

\bibitem{simclr}
Ting Chen, Simon Kornblith, Mohammad Norouzi, and Geoffrey~E. Hinton.
\newblock A simple framework for contrastive learning of visual
  representations.
\newblock In {\em Proceedings of the 37th International Conference on Machine
  Learning, {ICML} 2020, 13-18 July 2020, Virtual Event}, volume 119 of {\em
  Proceedings of Machine Learning Research}, pages 1597--1607. {PMLR}, 2020.

\bibitem{mocov2}
Xinlei Chen, Haoqi Fan, Ross~B. Girshick, and Kaiming He.
\newblock Improved baselines with momentum contrastive learning.
\newblock {\em CoRR}, abs/2003.04297, 2020.

\bibitem{pirl}
Ishan Misra and Laurens van~der Maaten.
\newblock Self-supervised learning of pretext-invariant representations.
\newblock In {\em 2020 {IEEE/CVF} Conference on Computer Vision and Pattern
  Recognition, {CVPR} 2020, Seattle, WA, USA, June 13-19, 2020}, pages
  6706--6716. {IEEE}, 2020.

\bibitem{3dcnn}
Shuiwang Ji, Wei Xu, Ming Yang, and Kai Yu.
\newblock 3d convolutional neural networks for human action recognition.
\newblock {\em {IEEE} Trans. Pattern Anal. Mach. Intell.}, 35(1):221--231,
  2013.

\end{thebibliography}
}

\begin{IEEEbiography}[{\includegraphics[width=1in,height=1.25in,clip,keepaspectratio]{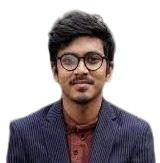}}]{Siladittya Manna}{\space}a received Dual Degree (B.Tech-M.Tech) in Electronics
and Telecommunication Engineering from Indian Institute of Engineering
Science and Technology, Shibpur, India in 2019. He is currently a Senior
Research Fellow at the Computer Vision and Pattern Recognition Unit,
Indian Statistical Institute, Kolkata, India. His research interests include
Self-supervised Learning, Computer Vision, Medical Image Analysis.

\end{IEEEbiography}

\begin{IEEEbiography}[{\includegraphics[width=1in,height=1.25in,clip,keepaspectratio]{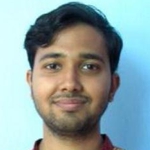}}]{Saumik Bhattacharya}{\space}is an assistant professor in the Department of Electronics and Electrical Communication Engineering, Indian Institute of Technology Kharagpur. He received the B.Tech. degree in Electronics and Communication Engineering from the West Bengal University of Technology, Kolkata in 2011, and the Ph.D. degree in Electrical Engineering from IIT Kanpur, Kanpur, India, in 2017. His research interests include image processing, computer vision, and machine learning.
\end{IEEEbiography}

\begin{IEEEbiography}[{\includegraphics[width=1in,height=1.25in,clip,keepaspectratio]{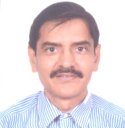}}]{Umapada Pal}{\space}received his Ph.D. in 1997 from Indian Statistical Institute, Kolkata, India. He did his Post Doctoral research at INRIA (Institut National de Recherche en Informatique et en Automatique), France. From January 1997, he is a Faculty member of Computer Vision and Pattern Recognition Unit of the Indian Statistical Institute, Kolkata and at present he is a Professor. He is a fellow of International Association of Pattern Recognition (IAPR). His fields of research interest include Digital Document Processing, Optical Character Recognition, Biometrics, Word spotting, Video Document Analysis, Computer vision etc.
\end{IEEEbiography}

\end{document}